\documentclass{article}

\PassOptionsToPackage{numbers, compress}{natbib}

\usepackage[final]{neurips_2024}

\usepackage[utf8]{inputenc} 
\usepackage[T1]{fontenc}    
\usepackage{setspace}
\usepackage{hyperref}       
\usepackage{url}            
\usepackage{booktabs}       
\usepackage{amsfonts}       
\usepackage{nicefrac}       
\usepackage{microtype}      
\usepackage{xcolor}         
\usepackage{graphicx}

\usepackage{multirow}
\usepackage{epsfig}
\usepackage{amsmath}
\usepackage{amssymb}
\usepackage{subfigure}
\usepackage{marvosym}
\usepackage{color}
\usepackage{threeparttable}
\usepackage{algorithm}
\usepackage{algpseudocode}
\usepackage{amsthm}
\usepackage{pifont}
\usepackage{transparent}

\title{AutoTimes: Autoregressive Time Series Forecasters via Large Language Models}

\author{
  Yong Liu\thanks{Equal Contribution},
  Guo Qin\footnotemark[1],
  Xiangdong Huang,
  Jianmin Wang,
  Mingsheng Long\textsuperscript{\Letter} \\
  School of Software, BNRist, Tsinghua University, China \\
  {\small\texttt{\{liuyong21,qinguo24\}@mails.tsinghua.edu.cn,}}\\
  {\small\texttt{\{huangxdong,jimwang,mingsheng\}@tsinghua.edu.cn}}
}

\begin{document}

\maketitle

\begin{abstract}
Foundation models of time series have not been fully developed due to the limited availability of time series corpora and the underexploration of scalable pre-training. Based on the similar sequential formulation of time series and natural language, increasing research demonstrates the feasibility of leveraging large language models (LLM) for time series. Nevertheless, the inherent autoregressive property and decoder-only architecture of LLMs have not been fully considered, resulting in insufficient utilization of LLM abilities. To fully revitalize the general-purpose token transition and multi-step generation capability of large language models, we propose \textbf{AutoTimes} to repurpose LLMs as \textbf{Auto}regressive \textbf{Time} \textbf{s}eries forecasters, which projects time series into the embedding space of language tokens and autoregressively generates future predictions with arbitrary lengths. Compatible with any decoder-only LLMs, the consequent forecaster exhibits the flexibility of the lookback length and scalability with larger LLMs. Further, we formulate time series as prompts, extending the context for prediction beyond the lookback window, termed \textbf{in-context forecasting}. By introducing LLM-embedded textual timestamps, AutoTimes can utilize chronological information to align multivariate time series. Empirically, AutoTimes achieves state-of-the-art with $0.1\%$ trainable parameters and over $5\times$ training/inference speedup compared to advanced LLM-based forecasters. Code is available at this repository: \url{https://github.com/thuml/AutoTimes}.
\end{abstract}

\section{Introduction}
\label{sec:intro}
Time series forecasting is of crucial demand in real-world applications, covering various domains including climate, economics, energy, operations, etc.~\cite{liu2023itransformer, wu2021autoformer}. The growing challenges of general-purpose forecasting, where one model is versatile to handle variable-length scenarios~\cite{liu2024timer, woo2024unified} and the prediction is necessarily instructed by auxiliary information in other modalities~\cite{woo2023pushing, xue2023promptcast}, underscore the demand for foundation models~\cite{bommasani2021opportunities} of time series, which are aimed to exhibit enhanced capabilities, including multi-step generation, zero-shot generalization~\cite{zhou2023one, gruver2023large}, in-context learning and multimodal utilization~\cite{jin2023time}, thereby expanding the scope of time series forecasting to a wider range of situations.

Nevertheless, the development of time series foundation models has been hampered by the limited availability of large-scale pre-training datasets and the technical uncertainty of scalable backbones. In contrast, rapid progress is witnessed in large language models (LLM), facilitated by extensive text corpora~\cite{zhu2015aligning}, available pre-trained models~\cite{touvron2023llama}, and well-established adaptation techniques~\cite{hu2021lora}. Notably, language and time series share basic commonalities in sequence modeling and generation by learned token transitions, presenting opportunities to adopt off-the-shelf LLMs for time series. 

\begin{figure}[tbp]
  \includegraphics[width=\columnwidth]{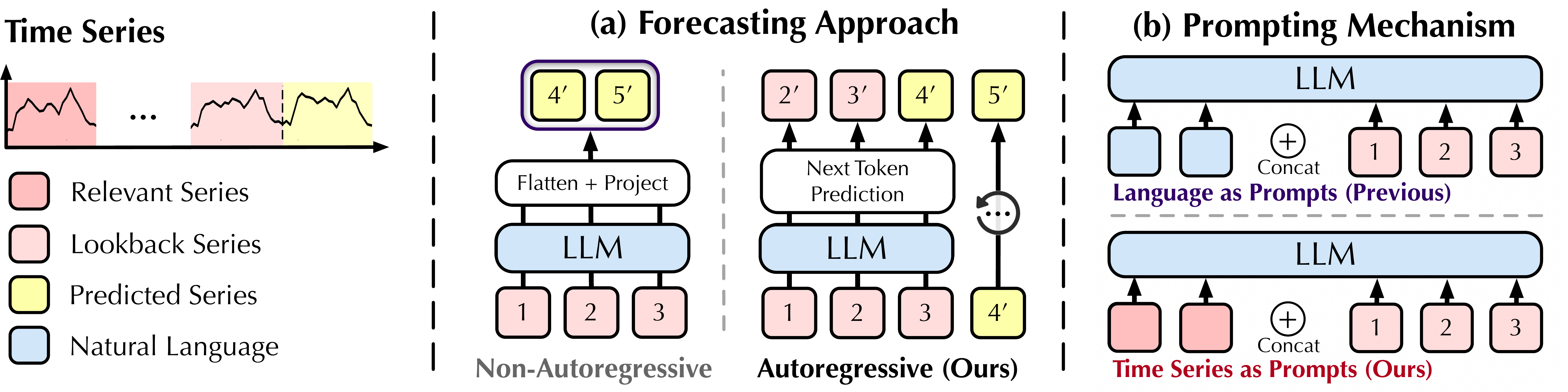}
  \centering
  \vspace{-18pt}
  \caption{(a) Prevalent LLM4TS methods non-autoregressively generate predictions with the globally flattened representation of lookback series, while large language models inherently predict the next tokens by autoregression~\cite{zhao2023survey}. (b) Previous methods adopt language prompts that may lead to the modality disparity, while we find time series can be self-prompted, termed \emph{in-context forecasting}.}
  \label{fig:motivation}
  \vspace{-12pt}
\end{figure}

Despite recent studies on large language models for time series (LLM4TS) achieving performance breakthroughs in current forecasting benchmarks~\cite{jin2023time}, the mechanism by which LLMs are aligned to the time series modality still remains obscure. The pilot work, FPT~\cite{zhou2023one} leverages LLMs as generic sequential representation extractors for time series, influencing subsequent LLM4TS methodologies. As depicted in Figure~\ref{fig:motivation} (a), the non-autoregressive approach, where time series are segmented into tokens, flattens and projects all lookback tokens for the prediction in a single step. However, it causes inconsistencies in both model structure and generative approach of LLMs: \emph{decoder-only models for autoregressive generation are converted to encoder-only and non-autoregressive forecasters}. 

Given that prior studies~\cite{dai2022can, wang2022language} reveal that generalization performance of LLMs is largely derived from the decoder-only structure trained autoregressively, talents of LLMs may not be fully exhibited. It is also supported by the recent rethinking of previous LLM4TS methods~\cite{tan2024language}, which generally lack the maintenance of autoregression, the essential characteristic of both large language models and statistical forecasters~\cite{box2015time, winters1960forecasting}. Therefore, autoregressive LLM4TS methods are underexplored, which can potentially unlock multi-step generation like LLMs, presenting one model for arbitrary lengths.

Motivated by the reflections, we propose \textbf{AutoTimes} to adapt LLMs as time series forecasters, which \emph{retrieves the consistency of autoregression with revitalized LLM capabilities to produce foundation models for time series forecasting}. Technically, we independently embed time series segments into the latent space of language models by the consistent training objective: next token prediction~\cite{bengio2000neural}. To fully leverage the inherent token transitions of LLMs and reduce the training cost, we freeze the LLM and establish token embedding and projection for time series, which only account for up to $0.1\%$ total parameters. The consequent forecaster adopts autoregressive inference like LLMs, which is no longer constrained to specific lookback/forecast lengths. Going beyond conventional time series forecasting, we propose \textbf{in-context forecasting} as shown in Figure~\ref{fig:motivation}, where time series can be self-prompted by relevant contexts. We further adopt LLM-embedded timestamps as the position embedding to utilize chronological information and align multiple variates. Our contributions are summarized as follows:

\begin{itemize}
    \item By refining the inconsistency of non-autoregressive LLM4TS methods, we propose to inherit the autoregressive property of LLMs, which frees our method from training respectively on the lookback length and allows arbitrary-length predictions with chronological awareness.
    \item We present AutoTimes, a simple but effective approach to acquire LLM-based forecasters by lightweight adaptation, which utilizes the inherent token transition as the future extrapolation of time series. Further, we propose in-context forecasting, which renovates the conventional paradigm by introducing relevant time series prompts to enhance forecasting.
    \item Compared with state-of-the-art methods, our repurposed forecaster achieves superior performance while saving over $80\%$ training and inference time, and further exhibits zero-shot generalizability, in-context forecasting, and scaling behavior empowered by LLMs.
\end{itemize}

\section{Related Work}
\label{sec:related}

\subsection{Autoregressive Models}
Autoregression is an essential concept of both language modeling and time series forecasting. Despite prevalent deep forecasters~\cite{das2023long, nie2022time, wu2022timesnet, zhou2021informer} adopt a non-autoregressive approach without the requirement of iterative forecasting, autoregression, the absent exploration in deep forecasters, serves as the fundamental principle of statistical methods, which enables variable-length predictions. The most well-known model, ARIMA~\cite{box2013box} is developed by incorporating differencing on AR and MA models, which are both autoregressive models with learned time-invariant transition from the past to the future. Incorporated with decomposition and pre-defined transitions, exponential smoothing~\cite{winters1960forecasting} and state space models (SSM)~\cite{durbin2012time, liu2023koopa} also take the same autoregressive formulation. 

Autoregressive language models~\cite{openai2023gpt, radford2019language} are trained with fine-grained supervision, where the generated token of each position is independently supervised. Consequently, they are not constrained by specific input/output lengths and excel at multi-step generation. Furthermore, existing LLMs are inherently autoregressive models~\cite{zhao2023survey}, which demonstrate advanced abilities that are not present in small models, such as the generalization~\cite{wang2022language}, scalability~\cite{brown2020language}, and task generality~\cite{radford2019language, raffel2020exploring}. Therefore, it is imperative to adapt off-the-shelf LLMs as autoregressive forecasters, which keeps the consistency to fully revitalize the model capacity and general-purpose token transitions.

\subsection{Large Language Models for Time Series}
With the immense advancement of large language model infrastructure, LLM4TS methods have been experiencing significant development in recent years. PromptCast~\cite{xue2023promptcast} reformulates time series as text pairs and accomplishes forecasting as a sentence-to-sentence task. LLMTime~\cite{gruver2023large} regards time series as numerical tokens, demonstrating the zero-shot generalizability in time series forecasting. FPT~\cite{zhou2023one} fine-tunes parameters of the LLM to adapt it as a general representation extractor serving for multiple time series analysis tasks. UniTime~\cite{liu2023unitime} adapts a language model across diverse time series for a unified forecaster of multiple domains. Based on thriving prompting techniques, deft language prompts~\cite{jin2023time, liu2023unitime} and soft prompting~\cite{cao2023tempo} for time series are further investigated.

LLM4TS methods have achieved performance breakthroughs in time series forecasting, but the cost of training and inference can sometimes be resource-consuming due to the immensity of LLMs. Recent revisiting of LLM4TS methods has revealed the inefficacy of LLMs adapted in the non-autoregressive approach~\cite{tan2024language}. By contrast, AutoTimes frozen LLMs, transfers the general-purpose token transition, and introduces minimal parameters to realize autoregressive next token prediction, thereby achieving better model efficiency and consistent utilization of large models. We further provide Table~\ref{tab:comparison} that categorizes prevalent LLM4TS methods by several essential aspects.

\subsection{Multimodal Language Models}
Multimodal models have been well-developed upon LLMs, among which vision language models (VLM) have experienced rapid growth~\cite{alayrac2022flamingo, openai2023gpt}. The booming pre-trained vision backbones~\cite{dosovitskiy2020image, radford2021learning}, together with the instruction tuning paradigm, has revealed the potential of LLMs for vision tasks, where visual tokens and language tokens are concatenated as the input of the LLM~\cite{li2022blip, liu2023visual}. Inspired by this, previous LLM4TS methods utilize instructive language tokens as prefix-prompts for time series analysis~\cite{jin2023time, sun2023test, xue2023promptcast}. Unlike previous works, our proposed method regards time series itself as the instructive prompt. It avoids the modality gap caused by concatenating time series and language tokens directly. We incorporate chronological information, the textual timestamp of time series, such that the language model can effectively perceive date and periodicity as the position embedding, and align simultaneous events from different time series~\cite{liu2023itransformer} for multivariate forecasting.

\begin{table}[htbp] %
  \caption{Comparison of LLM4TS methods: \emph{Autoregressive} categories LLM-based forecasters by whether to conduct autoregression. \emph{Freeze LLM} enables quick adaptation, which would otherwise require significant resources for fine-tuning. \emph{Multimodal} refers to the utilization of information from other modalities. Prior to AutoTimes, none of the LLM4TS methods achieved all three.}\label{tab:comparison}
  \vspace{-2pt}
  \centering
  \begin{small}
    \begin{threeparttable}
    \setlength{\tabcolsep}{2.5pt}
    \resizebox{\textwidth}{!}{\begin{tabular}{c|ccccccccc}
    \toprule
    Method & \textbf{AutoTimes} & TimeLLM~\cite{jin2023time} & UniTime~\cite{liu2023unitime}  & FPT~\cite{zhou2023one} & LLMTime~\cite{gruver2023large} & TEST~\cite{sun2023test} & TEMPO~\cite{cao2023tempo} & PromptCast~\cite{xue2023promptcast}\\ 
    \midrule
    Autoregressive          & \ding{51} & {\transparent{0.5} \ding{55}} & {\transparent{0.5} \ding{55}} & {\transparent{0.5} \ding{55}} & \ding{51} & {\transparent{0.5} \ding{55}} & {\transparent{0.5} \ding{55}} & {\transparent{0.5} \ding{55}} \\
    Freeze LLM       & \ding{51} & \ding{51} & {\transparent{0.5} \ding{55}} & {\transparent{0.5} \ding{55}} & \ding{51} & \ding{51} & {\transparent{0.5} \ding{55}} & \ding{51} \\
    Multimodal  & \ding{51} & \ding{51} & \ding{51} & {\transparent{0.5} \ding{55}}  & {\transparent{0.5} \ding{55}} & \ding{51} & \ding{51} & \ding{51} \\
    \bottomrule
    \end{tabular}}
  \end{threeparttable}
  \end{small}
  \vspace{-8pt}
\end{table}

\section{Method}

The proposed AutoTimes adapts large language models for multivariate time series forecasting. Given lookback observations $\mathbf{x}_{1:L}=\{\mathbf{x}_1,\ldots,\mathbf{x}_L\}\in\mathbb{R}^{L\times C}$ with $L$ time steps and $C$ variates, the objective is to predict the future $F$ time steps $\mathbf{x}_{L+1:L+F}=\{\mathbf{x}_{L+1},\ldots,\mathbf{x}_{L+F}\}\in\mathbb{R}^{F\times C}$. Besides, the textual timestamp $\mathbf{a}_t$ (e.g. $2016/07/05$ $00$:$00$:$00$), as the most common covariate, is adopted for prediction, which is aligned with time points $\mathbf{x}_t\in\mathbb{R}^{C}$ at time $t$. The task is to train an LLM-based forecaster $f$ that is able to predict with the (varying) lookback length $L$ for the (arbitrary) forecast length $F$ as:
\begin{equation}
f : (\mathbf{x}_{1:L}, \mathbf{a}_{1:L+F}) \mapsto \hat{\mathbf{x}}_{L+1:L+F}.
\end{equation}

\begin{figure}[tbp]
  \includegraphics[width=\columnwidth]{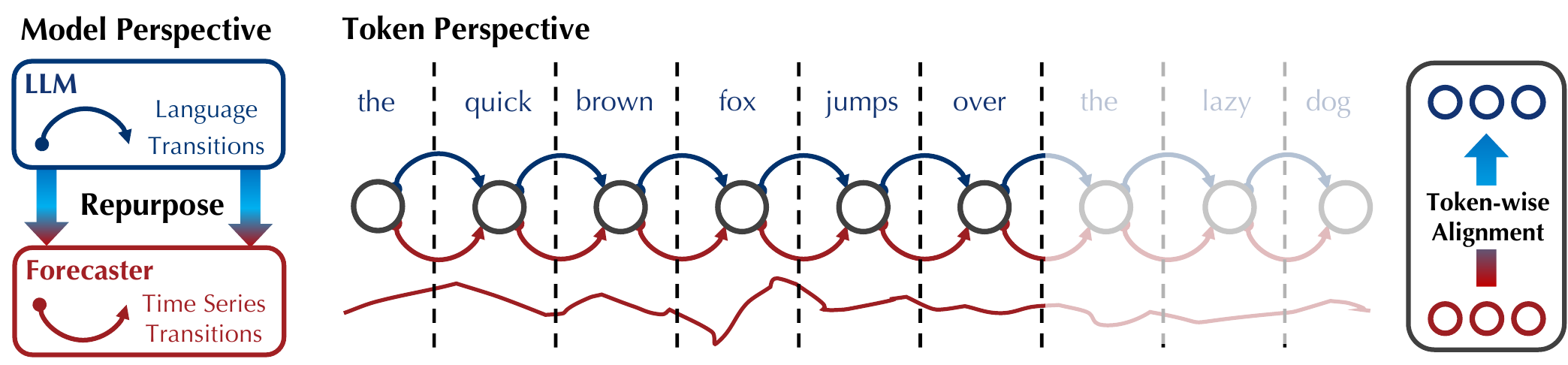}
  \centering
  \vspace{-15pt}
  \caption{An example to illustrate how AutoTimes adapts language models for time series forecasting.}
  \label{fig:mechanism}
  \vspace{-8pt}
\end{figure}

\subsection{Modality Alignment}

\paragraph{Time series token} To empower the forecaster with the capability to predict time series for arbitrary lengths, we repurpose autoregressive LLMs as time series forecasters as depicted in Figure~\ref{fig:mechanism}. Prior to this, we define time series token as the consecutive and non-overlapping segment of a single variate. It is regarded as the common token of the LLM-based forecaster, which encompasses series variations and mitigates excessively long autoregression. To focus on modeling temporal variations, our forecaster predicts each variate independently. Beyond Channel Independence~\cite{nie2022time} that implicitly captures the multivariate correlation~\cite{liu2023itransformer} by shared parameters, AutoTimes converts timestamps into position embeddings and explicitly aligns simultaneous segment tokens, which is detailed in the next paragraph. Therefore, we simplify $\mathbf{x}_t$ as the time point of specific variate $x_t\in\mathbb{R}$. Given a single-variate time series of context length $NS$, the $i$-th segment of length $S$ is denoted as:
\begin{equation}
\mathbf{s}_i=\{x_{(i-1)S+1},\dots,x_{iS}\}\in\mathbb{R}^{S},\ i=1,\dots,N.
\end{equation}
Considering the general-purpose token transition, we freeze the parameters of large language models. To realize the token-wise alignment between time series tokens and language tokens, we establish $\operatorname{SegmentEmbedding}(\cdot):\mathbb{R}^{S}\mapsto\mathbb{R}^{D}$ that independently embeds segments into the latent space:
\begin{equation}
\mathbf{SE}_i = \operatorname{SegmentEmbedding}(\mathbf{s}_i), \ i=1,\dots,N,
\end{equation}
where $D$ is consistent with the dimension of the LLM.

\paragraph{Position embedding} Timestamp, an essential covariate indicating the chronological information, is generally utilized as an extra embedding in previous deep forecasters~\cite{wu2021autoformer, zhou2021informer}. However, increasing models~\cite{das2023long, nie2022time, zeng2023transformers} have discarded the embedding and found the performance will not be greatly affected, implying the improper encoding of timestamps. In contrast, textual timestamps have been demonstrated as an enhancement in LLM4TS methods, which are always formulated into prefix-prompts~\cite{jin2023time, liu2023unitime}. Nevertheless, it also leads to excessive context length, impeding LLMs from paying sufficient attention to time series tokens and inducing time-consuming feed-forwarding. Inspired by the functionality of position embedding, which incorporates information about the relative or absolute position of the tokens~\cite{vaswani2017attention}. We adopt LLM-embedded timestamps as position embeddings to utilize temporal information and align simultaneous events (segments) from different varieties.

\begin{figure}[tbp]
  \includegraphics[width=\columnwidth]{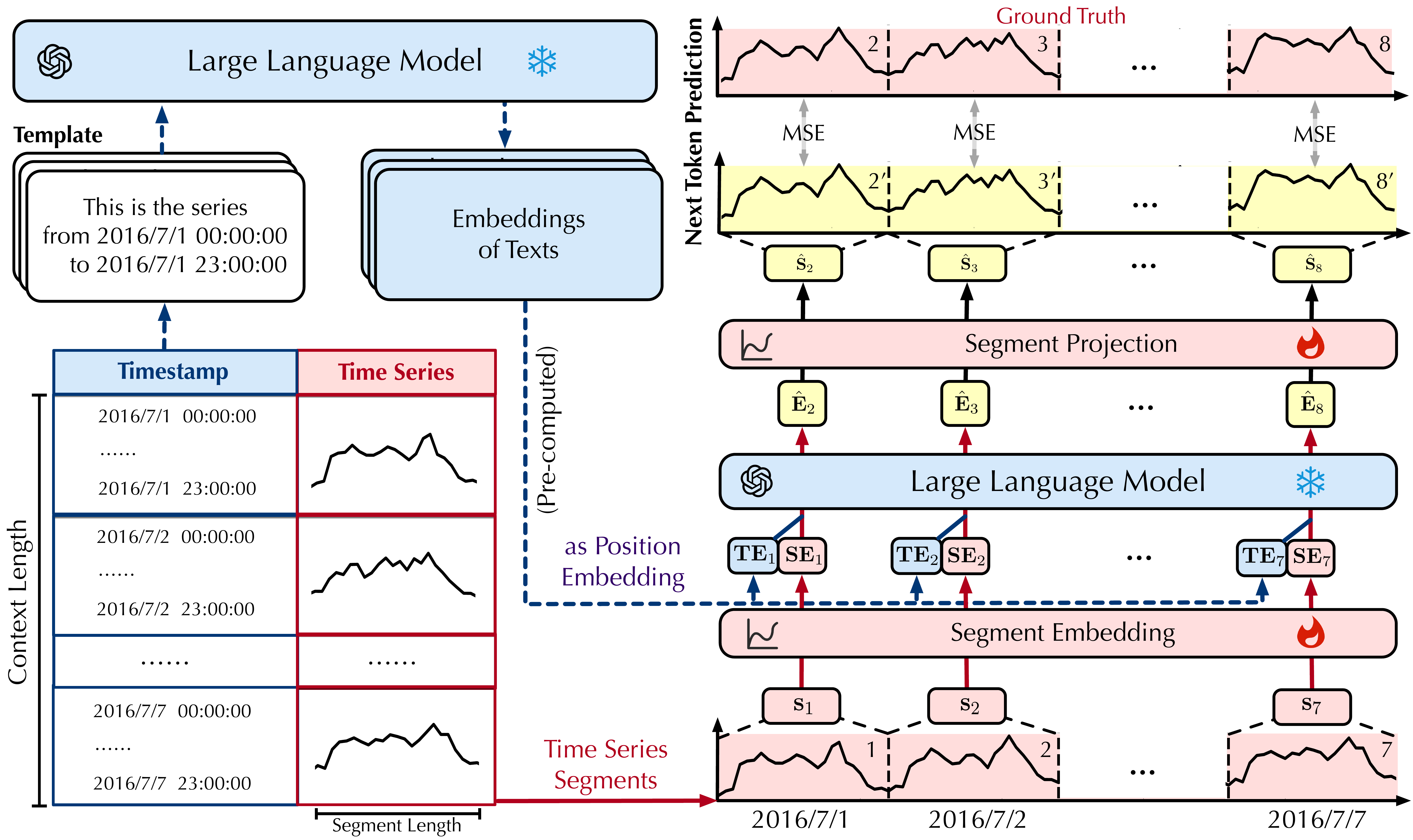}
  \centering
  \vspace{-15pt}
  \caption{Overview of AutoTimes: (1) time series and corresponding timestamps are segmented; (2) textual timestamps are converted into the position embeddings by the LLM; (3) time series segments are embedded and projected by next token prediction, where intermediate layers of LLM are frozen.}
  \label{fig:method}
  \vspace{-10pt}
\end{figure}

Technically, we formulate the starting and end timestamps of corresponding segments by the template demonstrated in Figure~\ref{fig:method}. Experimentally, we observe that the simple template without deft design can consistently boost the forecasting performance in Appendix~\ref{sec:timestamp}, aiding the LLM-based forecaster to comprehend the date and align different variates based on Channel Independence. Since all the previous language tokens are visible to the special ending token \verb|<EOS>| of a sentence, we adopt the embedding of \verb|<EOS>| as $\mathbf{TE}_i\in\mathbb{R}^{D}$ as the position embedding from textual timestamps:
\begin{equation}
\mathbf{TE}_i = \operatorname{SelectLast}\big(\operatorname{LLM}(\operatorname{TimestampTemplate}(\mathbf{s}_i))\big).
\end{equation}
Notably, $\mathbf{TE}_i$ is pre-computed by LLMs such that runtime forwarding for language tokens is not required during training. Given that the latent space of the LLM locates both time series tokens and language tokens, the position embedding can be integrated with the corresponding time span without increasing the context length. Concretely, the token embedding $\mathbf{E}_i\in\mathbb{R}^{D}$ is obtained by:
\begin{equation}
\mathbf{E}_i = \mathbf{SE}_i + \mathbf{TE}_i.
\end{equation}

\subsection{Next Token Prediction}
As shown in Figure~\ref{fig:method}, prevalent LLMs~\cite{brown2020language, touvron2023llama} are endowed with the capability of predicting the next token $\mathbf{s}_i$ based on the preceding tokens $\mathbf{s}_{<i}$. We reutilize LLMs in a fully consistent approach and generate prediction of arbitrary lengths iteratively. Given a time series of context length $NS$, the input series is segmented and embedded into $N$ token embeddings $\{\mathbf{E}_1, \dots, \mathbf{E}_N\}$. The training objective is to independently generate the next tokens $\{\hat{s}_2, \dots, \hat{s}_{N+1}\}$. We feed the token embeddings $\mathbf{E}_i$ into the intermediate layers of the LLM, which inherently parameterize token transitions:
\begin{equation}
\{\hat{\mathbf{E}}_2, \dots, \hat{\mathbf{E}}_{N+1}\} = \operatorname{LLMLayers}(\{\mathbf{E}_1, \dots, \mathbf{E}_N\}).
\end{equation}
We adopt $\operatorname{SegmentProjection}(\cdot):\mathbb{R}^{D}\mapsto\mathbb{R}^{S}$ to independently projects embeddings to segments:
\begin{equation}
\hat{\mathbf{s}}_i = \operatorname{SegmentProjection}(\hat{\mathbf{E}}_i), \ i=2,\dots,N+1.
\end{equation}
Finally, each predicted segment is supervised by the token-wise ground truth to optimize the parameters of embedding and projection layers, which are simply implemented by multi-layer perceptrons:
\begin{equation}\label{equ:loss}
\mathcal{L}_{\text{MSE}} = \frac{1}{NS} \sum ||\mathbf{s}_i-\hat{\mathbf{s}}_i||_2^2,\ i=2,\dots,N.
\end{equation}
Notably, the context length $NS$ is decided during training, representing the maximum input length during inference. Therefore, one consequent forecaster is suitable for different input lengths like the LLM, validated in Appendix~\ref{sec:vary_lookback}. Moreover, AutoTimes can generate predictions of arbitrary lengths by iterative multi-step forecasting, proven to overcome error accumulation better than state-of-the-art forecasters in Section~\ref{sec:one_for_all_forecast}, since autoregressive LLMs inherently excel at multi-step generation:
\begin{equation}
\hat{\mathbf{s}}_i = \operatorname{LLMForecaster}(\mathbf{s}_{<i}),\ i=1,\dots,\frac{F}{S}.
\end{equation}
Instead of respectively training models on different lookback/forecast lengths, AutoTimes handles all the scenarios by one model. Surprisingly, with the consistency of autoregression, it also inherits notable generalizability and scaling behavior of LLMs, which is demonstrated in Sections~\ref{sec:zeroshot} and~\ref{sec:analysis}.

\subsection{In-Context Forecasting}\label{sec:icl_forecast}
Large language models are capable of generating expected outputs based on provided task demonstrations from downstream datasets without gradient updating, known as the in-context learning ability. The task demonstrations are generally constituted by paired questions and answers~\cite{zhao2023survey}. Formally, the context $\mathcal{C} = \{g(x^{(1)}, y^{(1)}),\dots, g(x^{(m)}, y^{(m)})\}$ represents a set of demonstrations with $m$ pairs, where $g(\cdot)$ is the template that transforms each question and answer into natural language. 

In terms of time series forecasting, we propose to constitute the pair by lookback-forecast windows, which are exactly represented as successive time points from earlier historical observations. Hence, we use time series in target datasets as prompts, \emph{extending the context for prediction beyond consecutive lookback series}. We denote the extended context as $\mathcal{C}$, which contains $m$ time series prompts $\operatorname{tsp}^{(j)}$:
\begin{equation}\label{equ:tsp}
\mathcal{C}=\{\operatorname{tsp}^{(j)}=\mathbf{x}_{\le t_j}|\text{ earlier historical time series}\},\ j=1,\dots,m,\ t_j \le L.
\end{equation}

During training, we first obtain an LLM-based forecaster on a source dataset and select time series prompts from the downstream target dataset based on a unified strategy. During inference, we ensure all the prompts appear before the window to be predicted, such that there is no data leakage from future information. As shown in Figure~\ref{fig:icl_forecast}, we concatenate time series prompts with lookback series and feed them as the context of the forecaster, termed \emph{in-context forecasting}:
\begin{equation}
f : (\mathcal{C}, \mathbf{x}_{1:L}, \mathbf{a}_{1:L+F}) \mapsto \hat{\mathbf{x}}_{L+1:L+F}.
\end{equation}

\section{Experiments}
\label{sec:exp}

We conduct thorough evaluations of the performance of AutoTimes, including time series forecasting, zero-shot forecasting, and the proposed in-context forecasting. Additional analyses are included to evaluate the generality, scaling behavior, and adaptation cost of large language models. Detailed code implementation for reproduction is provided in our public code repository.

\subsection{Time Series Forecasting}\label{sec:one_for_all_forecast}
\paragraph{Benchmarks} For long-term time series forecasting, we extensively include real-world datasets, including ETTh1, ECL, Traffic, Weather~\cite{wu2021autoformer}, and Solar-Energy~\cite{liu2023itransformer}. For short-term forecasting, we adopt the well-acknowledged M4 competition~\cite{makridakis2020m4}. Detailed descriptions are provided in Appendix~\ref{sec:datasets}.

\paragraph{Baselines} We compare AutoTimes with state-of-the-art models, including advanced LLM4TS methods: TimeLLM~\cite{jin2023time}, UniTime~\cite{liu2023unitime}, and FPT~\cite{zhou2023one}; well-acknowledged deep forecasters: iTransformer~\cite{liu2023itransformer}, DLinear~\cite{zeng2023transformers}, PatchTST~\cite{nie2022time}, and TimesNet~\cite{wu2022timesnet}. For the challenging short-term forecasting, we further include competitive baselines: Koopa~\cite{liu2023koopa}, N-HiTS~\cite{challu2023nhits} and N-BEATS~\cite{oreshkin2019n}. All baselines are officially implemented or reported. We adopt LLaMA-7B~\cite{touvron2023llama} as our base LLM. Detailed implementations, error bars, and hyperparameter analysis are provided in Appendix~\ref{sec:implementation} and~\ref{sec:hyperparams}.

\paragraph{Setups} For short-term forecasting, we follow the well-acknowledged TimesNet~\cite{wu2022timesnet}, which assesses the fundamental ability of forecasters in modeling temporal variations. For long-term forecasting, we establish a novel \emph{one-for-all} benchmark: a single forecaster is trained on one dataset and subsequently utilized for all prediction lengths. We highlight that this approach evaluates the basic versatility as foundation models of time series, which aims to break the prevailing practice of extensive training across diverse real-world scenarios. To be specific, we evaluate all methods by rolling forecasting: a model is trained with predetermined input/output lengths, and the predicted values are integrated as part of the input in subsequent iterations until reaching the desired forecast length. Therefore, the key to success in this task lies in mitigating multi-step error accumulation. Still, the conventional \emph{one-for-one} approach that trains forecasters respectively on each length is also provided in Table~\ref{tab:one_for_one}.

\begin{table}[htbp]
  \caption{Average short-term forecasting results on the M4~\cite{makridakis2020m4}. Full results are provided in Table~\ref{tab:forecast_short_full}.}
  \vspace{-10pt}
  \centering
  \begin{threeparttable}
  \begin{small}
  \label{tab:forecast_short}
  \renewcommand{\multirowsetup}{\centering}
  \setlength{\tabcolsep}{2.9pt}
  \resizebox{\textwidth}{!}{\begin{tabular}{c|r|ccccccccccccccc}
    \toprule
    \multicolumn{2}{c|}{\scalebox{0.85}{Models}} & 
    \multicolumn{1}{c}{\rotatebox{0}{\scalebox{0.85}{\textbf{AutoTimes}}}} &
    \multicolumn{1}{c}{\rotatebox{0}{\scalebox{0.85}{TimeLLM}}} &
    \multicolumn{1}{c}{\rotatebox{0}{\scalebox{0.85}{FPT}}} &
    \multicolumn{1}{c}{\rotatebox{0}{\scalebox{0.85}{Koopa}}} &
    \multicolumn{1}{c}{\rotatebox{0}{\scalebox{0.85}{N-HiTS}}} &
    \multicolumn{1}{c}{\rotatebox{0}{\scalebox{0.85}{DLinear}}} &
    \multicolumn{1}{c}{\rotatebox{0}{\scalebox{0.85}{PatchTST}}} &
    \multicolumn{1}{c}{\rotatebox{0}{\scalebox{0.85}{TimesNet}}} &
    \multicolumn{1}{c}{\rotatebox{0}{\scalebox{0.85}{FiLM}}} &
    \multicolumn{1}{c}{\rotatebox{0}{\scalebox{0.85}{{N-BEATS}}}} \\ 
    \toprule
    \multirow{3}{*}{\rotatebox{90}{\scalebox{0.85}{Average}}}
    & \scalebox{0.85}{sMAPE} & \textbf{11.831} & 11.983 & 11.991 & \underline{11.863} & 11.960  & 12.418 & 13.022 & 11.930 & 12.489 & 11.910  \\
    & \scalebox{0.85}{MASE}  & \textbf{1.585} & \underline{1.595} & 1.600 & \underline{1.595}  & 1.606 & 1.656   & 1.814  & 1.597    & 1.690  & 1.613\\
    & \scalebox{0.85}{OWA}   & \textbf{0.850} & 0.859 & 0.861 &  \underline{0.858}  & 0.861  & 0.891  & 0.954  & 0.867   & 0.902  & 0.862\\
    \bottomrule
  \end{tabular}}
  \end{small}
  \end{threeparttable}
  \vspace{-5pt}
\end{table}

\begin{table}[htbp]
  \caption{Long-term forecasting results of one-for-all: we conduct rolling forecasting with a single model trained on each dataset and accomplish four desired forecast lengths in $\{96, 192, 336, 720\}$. AutoTimes adapt LLMs with the context length $C=672$. We set the input length $L=672$ and output length $F=96$ in other methods. All results are averaged. Full results is provided in Table~\ref{tab:one_model_full}.}
  \label{tab:one_model}
  \centering
  \begin{threeparttable}
  \begin{small}
  \renewcommand{\multirowsetup}{\centering}
  \setlength{\tabcolsep}{2pt}
  \resizebox{\textwidth}{!}{\begin{tabular}{c|c|cccccccccccccccc}
    \toprule
    \multicolumn{2}{c|}{Models} & \multicolumn{2}{c}{\scalebox{0.9}{\textbf{AutoTimes} }} &
    \multicolumn{2}{c}{\scalebox{0.80}{TimeLLM~\cite{jin2023time}}}  & 
    \multicolumn{2}{c}{\scalebox{0.80}{UniTime~\cite{liu2023unitime}}} &
    \multicolumn{2}{c}{\scalebox{0.90}{FPT~\cite{zhou2023one}}}  & 
    \multicolumn{2}{c}{\scalebox{0.80}{iTrans.~\cite{liu2023itransformer}}} &
    \multicolumn{2}{c}{\scalebox{0.80}{DLinear~\cite{zeng2023transformers}}} & 
    \multicolumn{2}{c}{\scalebox{0.75}{PatchTST~\cite{nie2022time}}} & 
 \multicolumn{2}{c}{\scalebox{0.75}{TimesNet~\cite{wu2022timesnet}}}  \\
     \cmidrule(lr){1-2} \cmidrule(lr){3-4} \cmidrule(lr){5-6}\cmidrule(lr){7-8} \cmidrule(lr){9-10}\cmidrule(lr){11-12}\cmidrule(lr){13-14}\cmidrule(lr){15-16}\cmidrule(lr){17-18}
    \multicolumn{2}{c|}{Metric} & \scalebox{0.90}{MSE} & \scalebox{0.90}{MAE} & \scalebox{0.90}{MSE} & \scalebox{0.90}{MAE} & \scalebox{0.90}{MSE} & \scalebox{0.90}{MAE} & \scalebox{0.90}{MSE} & \scalebox{0.90}{MAE} & \scalebox{0.90}{MSE} & \scalebox{0.90}{MAE} & \scalebox{0.90}{MSE} & \scalebox{0.90}{MAE} & \scalebox{0.90}{MSE} & \scalebox{0.90}{MAE} & \scalebox{0.90}{MSE} & \scalebox{0.90}{MAE}  \\
    \toprule
    \multicolumn{2}{c|}{\scalebox{0.90}{ETTh1}}
    & \scalebox{0.90}{\textbf{0.389}} & \scalebox{0.90}{\textbf{0.422}} 
    & \scalebox{0.90}{0.412} & \scalebox{0.90}{0.437} 
    & \scalebox{0.90}{0.683} & \scalebox{0.90}{0.596}    
    & \scalebox{0.90}{0.429} & \scalebox{0.90}{0.439} 
    & \scalebox{0.90}{0.421} & \scalebox{0.90}{0.445}
    & \scalebox{0.90}{0.426} & \scalebox{0.90}{0.444} 
    & \scalebox{0.90}{\underline{0.409}} & \scalebox{0.90}{\underline{0.430}} 
    & \scalebox{0.90}{0.495} & \scalebox{0.90}{0.491} \\
    
    \midrule
    \multicolumn{2}{c|}{\scalebox{0.90}{ECL}} 
    & \scalebox{0.90}{\textbf{0.159}} & \scalebox{0.90}{\textbf{0.253}} 
    & \scalebox{0.90}{0.181} & \scalebox{0.90}{0.288}
    & \scalebox{0.90}{0.325} & \scalebox{0.90}{0.399} 
    & \scalebox{0.90}{0.184} & \scalebox{0.90}{0.284}
    & \scalebox{0.90}{\underline{0.164}} & \scalebox{0.90}{\underline{0.258}}
    & \scalebox{0.90}{0.165} & \scalebox{0.90}{0.265} 
    & \scalebox{0.90}{0.169} & \scalebox{0.90}{0.268} 
    & \scalebox{0.90}{0.201} & \scalebox{0.90}{0.303} \\

    \midrule
    \multicolumn{2}{c|}{\scalebox{0.90}{Weather}} 
    & \scalebox{0.90}{0.235} & \scalebox{0.90}{0.273} 
    & \scalebox{0.90}{\textbf{0.225}} & \scalebox{0.90}{\textbf{0.266}} 
    & \scalebox{0.90}{0.461} & \scalebox{0.90}{0.459} 
    & \scalebox{0.90}{0.228} & \scalebox{0.90}{\textbf{0.266}}  
    & \scalebox{0.90}{0.266} & \scalebox{0.90}{0.291} 
    & \scalebox{0.90}{0.239} & \scalebox{0.90}{0.291} 
    & \scalebox{0.90}{\underline{0.226}} & \scalebox{0.90}{\underline{0.268}} 
    & \scalebox{0.90}{0.264} & \scalebox{0.90}{0.293} \\
    
    \midrule
    \multicolumn{2}{c|}{\scalebox{0.90}{Traffic}} 
    & \scalebox{0.90}{\textbf{0.374}} & \scalebox{0.90}{\textbf{0.264}} 
    & \scalebox{0.90}{0.410} & \scalebox{0.90}{0.303} 
    & \scalebox{0.90}{0.584} & \scalebox{0.90}{0.367} 
    & \scalebox{0.90}{0.461} & \scalebox{0.90}{0.326}
    & \scalebox{0.90}{\underline{0.384}} & \scalebox{0.90}{\underline{0.274}}
    & \scalebox{0.90}{0.423} & \scalebox{0.90}{0.298} 
    & \scalebox{0.90}{0.391} & \scalebox{0.90}{0.275} 
    & \scalebox{0.90}{0.602} & \scalebox{0.90}{0.322} \\
    
    \midrule
    \multicolumn{2}{c|}{\scalebox{0.90}{Solar.}} 
    & \scalebox{0.90}{\textbf{0.197}} & \scalebox{0.90}{\textbf{0.242}}  
    &  \scalebox{0.90}{0.263} & \scalebox{0.90}{0.335}
    & \scalebox{0.90}{0.392} & \scalebox{0.90}{0.462}
    & \scalebox{0.90}{0.236} & \scalebox{0.90}{0.303}
    & \scalebox{0.90}{0.213} & \scalebox{0.90}{0.291}  
    & \scalebox{0.90}{0.222} & \scalebox{0.90}{0.283} 
    & \scalebox{0.90}{\underline{0.202}} & \scalebox{0.90}{\underline{0.269}} 
    & \scalebox{0.90}{0.213} & \scalebox{0.90}{0.295} \\
    \bottomrule
  \end{tabular}}
  \end{small}
  \end{threeparttable}
  \vspace{-5pt}
\end{table}

\paragraph{Results} The average results are presented in Table~\ref{tab:forecast_short}-~\ref{tab:one_model}, with the best results in \textbf{bold} and the second best \underline{underlined}. AutoTimes consistently outperforms all counterparts of short-term forecasting in Table~\ref{tab:forecast_short}, demonstrating the basic ability of LLM-based forecasters to capture diverse series variations. Further, in the one-for-all long-term scenarios, AutoTimes surpasses other LLM4TS methods and deep forecasters in $80\%$ datasets in Table~\ref{tab:one_model}, outperforming previous state-of-the-art TimeLLM by $9.12\%$ in average. Compared with other forecasters trained in the one-for-one scenario in Table~\ref{tab:one_for_one}, AutoTimes still achieved state-of-the-art performance in $70\%$ of settings without respective training. 

By diving into the proposed one-for-all and the traditional one-for-one benchmarks in Table~\ref{tab:one_model} and~\ref{tab:one_for_one}, it is notable that prevalent deep forecasters, such as Transformer-based forecasters and DLinear, can achieve competitive and even better results under rolling forecasting. Nevertheless, the performance of non-autoregressive LLM4TS methods can degenerate a lot without respective training. Therefore, it highlights our persistent utilization of autoregression and thorough leveraging of inherent token transitions of LLMs, thereby mitigating error accumulation during multi-step rolling forecasting.

\subsection{Zero-Shot Forecasting}\label{sec:zeroshot}

\paragraph{Setups} Large language models have exhibited remarkable zero-shot generalization capability~\cite{brown2020language}. To verify whether our LLM-based forecaster inherits this ability, where no training sample of the target domain is available, we assess the performance of zero-shot forecasting. Concretely, we adhere to the benchmark established by FPT~\cite{zhou2023one}, where the forecaster is initially trained on a source domain and subsequently evaluated on an unseen target domain. We conduct the transfer learning between the M3 and M4 competitions, both of which encompass abundant temporal variation patterns but follow different data distributions. We compare AutoTimes with deep forecasters and FPT as the only LLM4TS method, given that only FPT has exhibited zero-shot generalization in this benchmark.

\begin{table}[htbp]
  \caption{Zero-shot forecasting results in averaged SMAPE. M4 $\to$ M3 trains forecasters on the datasets of M4 and evaluates on M3, and vice versa. Detailed results are provided in Appendix~\ref{appendix:zero_shot}}
  \vspace{-10pt}
  \centering
  \begin{threeparttable}
  \begin{small}
  \label{tab:forecast_zeroshot}
  \renewcommand{\multirowsetup}{\centering}
  \setlength{\tabcolsep}{2.9pt}
    \resizebox{\textwidth}{!}{\begin{tabular}{c|ccccccccc}
    \toprule
    \multicolumn{1}{c|}{\scalebox{0.85}{Models}} & 
    \multicolumn{1}{c}{\rotatebox{0}{\scalebox{0.85}{\textbf{AutoTimes}}}} &
    \multicolumn{1}{c}{\rotatebox{0}{\scalebox{0.85}{FPT}}} &
    \multicolumn{1}{c}{\rotatebox{0}{\scalebox{0.85}{DLinear}}} &
    \multicolumn{1}{c}{\rotatebox{0}{\scalebox{0.85}{PatchTST}}} &
    \multicolumn{1}{c}{\rotatebox{0}{\scalebox{0.85}{TimesNet}}} &
    \multicolumn{1}{c}{\rotatebox{0}{\scalebox{0.85}{{NSFormer}}}} &
    \multicolumn{1}{c}{\rotatebox{0}{\scalebox{0.85}{FEDFormer}}} &
    \multicolumn{1}{c}{\rotatebox{0}{\scalebox{0.85}{Informer}}} &
    \multicolumn{1}{c}{\rotatebox{0}{\scalebox{0.85}{Reformer}}} \\
    \toprule
    \scalebox{0.85}{M4$\ \to\ $M3} & \textbf{12.75} & \underline{13.06} & 14.03 & \underline{13.06} & 14.17 & 15.29 & 13.53 & 15.82 & 13.37  \\
    \midrule
    \scalebox{0.85}{M3$\ \to\ $M4}  & \textbf{13.036} & \underline{13.125} & 15.337 & 13.228  & 14.553  & 14.327  & 15.047  & 19.047  & 14.092\\
    \bottomrule
  \end{tabular}}
  \end{small}
  \end{threeparttable}
\end{table}

\paragraph{Results} The comprehensive results of zero-shot forecasting are presented in Table~\ref{tab:forecast_zeroshot}. AutoTimes demonstrates superior performance compared to deep forecasters and FPT in both M4 $\to$ M3 and M3 $\to$ M4 scenarios. It is evident that LLM4TS methods generally achieve improved performance in this task due to the enhanced model capacity, leading to a 15\% SMAPE reduction compared with the efficient forecaster DLinear. Despite sharing the same Transformer backbone, LLM4TS methods still outperform PatchTST due to the transferable knowledge pre-trained on large corpora of sequences. This underscores the advantage of leveraging LLMs for time series forecasting. Moreover, AutoTimes inherits general-purpose token transitions, surpassing FPT without tuning intermediate LLM layers.

\subsection{In-Context Forecasting}\label{sec:incontext}
\begin{figure}[tbp]
  \includegraphics[width=\columnwidth]{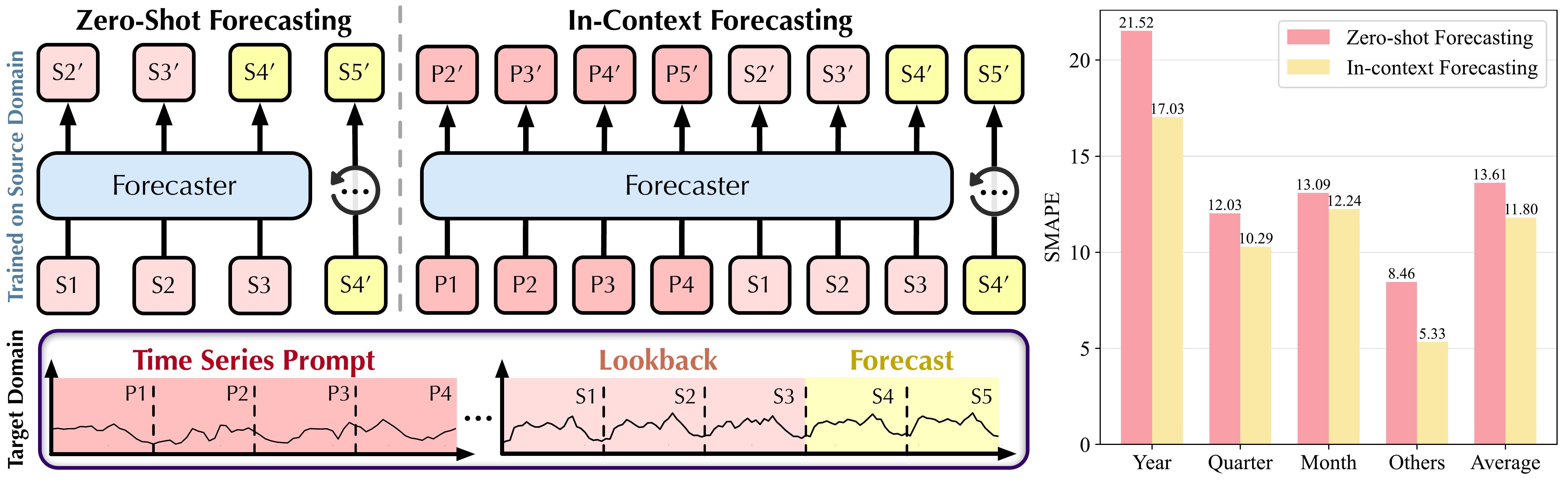}
  \centering
  \vspace{-15pt}
	\caption{Demonstration of in-context forecasting and results compared with zero-shot. We uniformly select the foremost time points from the target domain as prompts and concatenate them with lookback to obtain the prediction. AutoTimes adapts LLMs on the source domain with a larger context length to place the additional time series prompt. Supplementary showcases are provided in Figure~\ref{fig:icl_example}.}
	\label{fig:icl_forecast}
 \vspace{-8pt}
\end{figure}

\paragraph{Setups} We conduct in-context forecasting on AutoTimes, which is depicted in Figure~\ref{fig:icl_forecast}. Similar to zero-shot forecasting, the task is to apply a forecaster, trained on a source dataset, to an unseen target dataset. Additionally, several task demonstrations from the target domain, referred to as time series prompts in Equation~\ref{equ:tsp}, are available during inference. Specifically, we concatenate these prompts with the lookback window to form the context for prediction. 

We adopt the aforementioned M4 $\to$ M3 scenario. Since the samples of the M3 dataset are univariate time series with different lengths, we always predict the last $F$ time points of each sample during inference. We set the lookback length $L=F$, and thus the length of time series prompts is $2F$. We set the number of prompts $m=1$. Therefore, we initially train an LLM-based forecaster on the source M4 dataset with the context length of $3F$. We adopt an intuitive strategy to select the prompt: uniformly adopting the first $2F$ time points of the time series as the corresponding prompt. Supposing the lookback series starts after time $t\ (\ge F)$, in-context forecasting is formulated as:
\begin{equation} 
f : (\{x_{1:2F}\}, x_{t+1:t+F}, \mathbf{a}_{t+1:t+2F}) \mapsto \hat{x}_{t+F+1,t+2F}.
\end{equation}
To prevent data leakage of future information, too short samples are discarded to prevent the overlap between the prompt and the future prediction. The implementation details of in-context forecasting are provided in Appendix~\ref{sec:icl_detail}. We further investigate different prompt retrieval strategies. Insightful results are provided to reveal the influence of using time series prompts for interactive prediction and take-away instructions of prompt engineering in the time series modality.

\paragraph{Results} The quantitative results of in-context forecasting are provided on the right of Figure~\ref{fig:icl_forecast}. The results of zero-shot forecasting, where no downstream demonstration is available, are compared as the baseline. Benefiting from the time series prompts of the target domain, our LLM-based forecaster with the proposed in-context forecasting paradigm achieves consistent promotions on all M3 subsets and the averaged $13.3\%$ SMAPE reduction compared with zero-shot forecasting. In contrast to previous LLM4TS methods that rely on deft language prompts, LLMs adopted by AutoTimes can be instructed by time series itself with our intuitive prompting engineering. From the perspective of the forecasting paradigm, we extend the prediction context beyond the lookback window. To inherit token transitions of language models parameterized by intermediate layers, AutoTimes takes a crucial step by establishing a mapping between time series segments and the latent space of language tokens, which is however absent in non-autoregressive LLM4TS methods. Therefore, ensuring autoregression consistency enhances the effective utilization of LLMs as foundation models.

\subsection{Method Analysis}\label{sec:analysis}

\paragraph{Generality} Previous LLM4TS~\cite{jin2023time,  zhou2023one} methods focus on applying their approach to specific LLMs. We demonstrate that AutoTimes is compatible with any decoder-only LLMs. By extensively training LLM-based forecasters by AutoTimes based on prevalent LLMs, including GPT-2~\cite{radford2019language}, OPT~\cite{zhang2022opt}, and LLaMA~\cite{touvron2023llama}, we present the results in Table~\ref{tab:llm_ablation}, highlighting the generality of AutoTimes.

\begin{table}[ht]
  \caption{Averaged results of alternative language models. Full results are provided in Table~\ref{tab:llm_ablation_full}.}
  \vspace{-3pt}
  \label{tab:llm_ablation}
  \centering
  \begin{threeparttable}
  \begin{small}
  \renewcommand{\multirowsetup}{\centering}
  \resizebox{\textwidth}{!}{\begin{tabular}{c|cccccccccccc}
    \toprule
    LLM & \multicolumn{2}{c}{GPT-2 (124M)} &\multicolumn{2}{c}{OPT-350M} &\multicolumn{2}{c}{OPT-1.3B} &\multicolumn{2}{c}{OPT-2.7B} & \multicolumn{2}{c}{OPT-6.7B} &  \multicolumn{2}{c}{LLaMA-7B}  \\
    \cmidrule(lr){1-1} \cmidrule(lr){2-3} \cmidrule(lr){4-5} \cmidrule(lr){6-7}\cmidrule(lr){8-9}\cmidrule(lr){10-11}\cmidrule(lr){12-13}
    Metric & MSE & MAE & MSE & MAE & MSE & MAE & MSE & MAE & MSE & MAE & MSE & MAE\\
    \toprule
    ECL &   0.173 & 0.266 & 0.168 & 0.263  & 0.164 & 0.258  & 0.164 & 0.258 & 0.162 & 0.256 & 0.159 & 0.253   \\
    \midrule
    ETTh1 &   0.397 & 0.425 & 0.401 & 0.429 & 0.396 & 0.424 & 0.394 & 0.424 & 0.394 & 0.424 & 0.389 & 0.423\\
    \midrule
    Traffic &  0.406 & 0.276 & 0.405 & 0.277 & 0.397 & 0.271  & 0.394 & 0.269 & 0.393 & 0.270 & 0.374 & 0.264\\
    \midrule
    Weather & 0.242 & 0.278 & 0.240 & 0.275 & 0.240 & 0.276 & 0.243 & 0.277 & 0.247 & 0.282 & 0.235 & 0.273\\
    \bottomrule
  \end{tabular}}
  \vspace{-8pt}
  \end{small}
  \end{threeparttable}
\end{table}

\begin{figure}[htbp]
\begin{center}
    \centerline{\includegraphics[width=\textwidth]{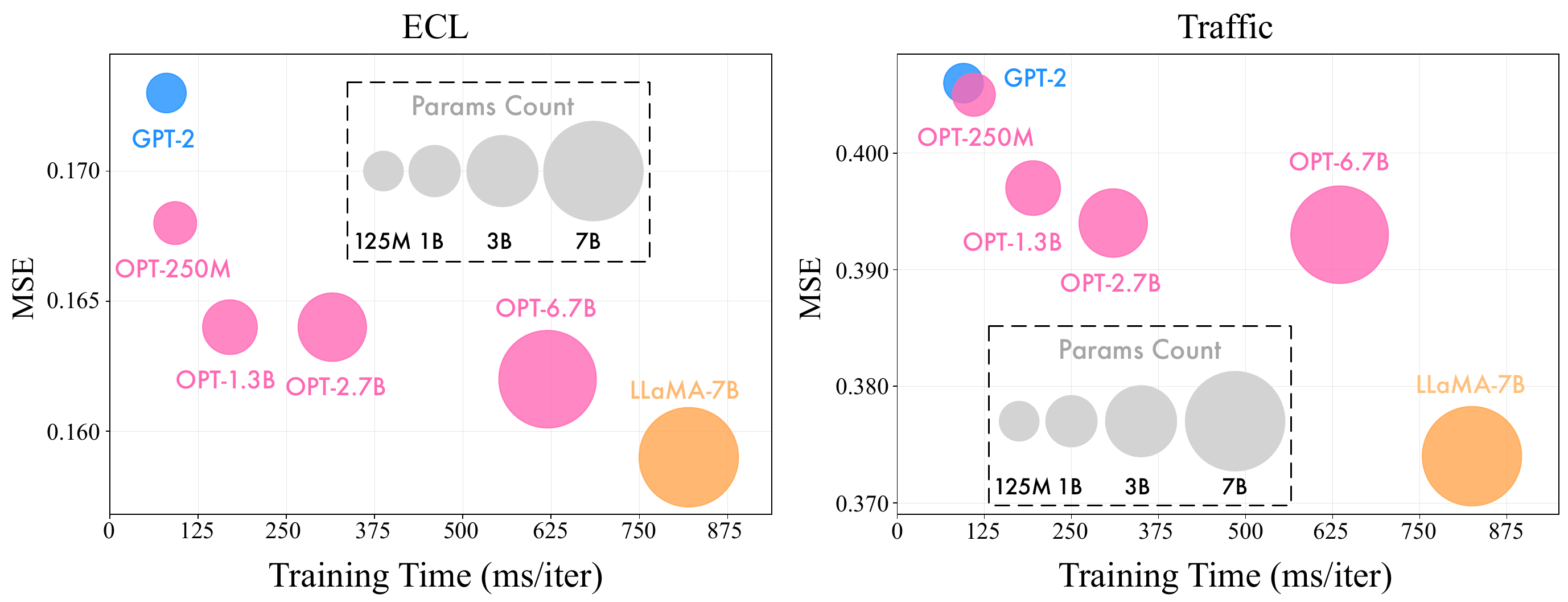}}
    \vspace{-5pt}
	\caption{Efficiency comparison of alternative LLMs, evaluated by the same configuration of Table~\ref{tab:llm_ablation}.}
	\label{fig:scalability}
\end{center}
\vspace{-25pt}
\end{figure}

\paragraph{Scaling behavior} Scalability is an essential characteristic that emerges from small models to large foundation models. By investigating the results presented in Table~\ref{tab:llm_ablation}, we observe that the prediction accuracy of the forecaster generally improves with the increase in LLM parameters. This scaling behavior of LLM-based forecasters introduces a trade-off between performance and adaptation cost. To provide a comprehensive assessment, we evaluate each adapted forecaster from three perspectives: performance, training speed, and parameters, as presented in Figure~\ref{fig:scalability}. We observe that the largest LLaMA-7B consistently delivers optimal forecasting performance. As a relatively small language model, OPT-1.3B exhibits good parameter efficiency as an out-of-the-box forecaster.

\begin{figure}[ht]
\vspace{2pt}
\begin{center}
\vspace{-5pt}
    \centerline{\includegraphics[width=\textwidth]{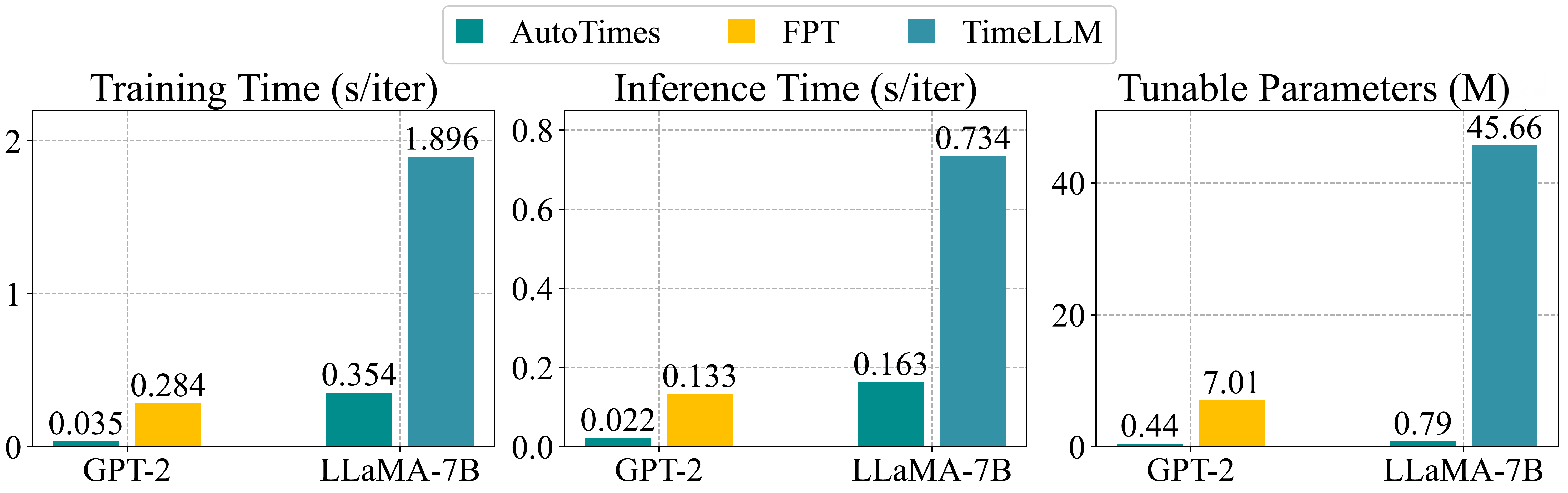}}
    \vspace{-5pt}
	\caption{Comparison of AutoTimes and other LLM4TS methods in terms of training/inference time and tunable parameters with the same batch size ($224$) on the ETTh1 dataset.}
	\label{fig:efficiency}
\end{center}
\vspace{-25pt}
\end{figure}

\paragraph{Adaptation cost} To mitigate the substantial cost of adapting large language models, AutoTimes introduces minimal parameters with all intermediate layers of LLM frozen. Additionally, we seamlessly integrate the language tokens (e.g. textual timestamps) without excessive context length and runtime overhead for training, thereby significantly reducing the adaptation cost. Figure~\ref{fig:efficiency} presents a comprehensive efficiency analysis with advanced LLM4TS methods: FPT is applicable on GPT-2 and TimeLLM is applicable on LLaMA-7B. Not only does AutoTime achieve better results in Table~\ref{tab:one_model}, but its training and reasoning time is also greatly reduced, bringing over 5$\times$ speedup on average. In terms of parameter efficiency, AutoTimes focuses on establishing the embedding for time series segments, which is simply implemented by the MLP ($0.79M$) account for $0.1\%$ parameters of the LLM ($7B$). Therefore, the results affirm the effectiveness of reutilizing the inherent token transition.

\paragraph{Ablation study} Recent research has raised doubts about the validity of previous LLM4TS methods~\cite{tan2024language}, which predominantly adopt non-autoregression, that is, treating the LLM as a BERT-style pre-trained backbone and utilize a globally flattening projector on all lookback tokens. Here, we provide a thorough ablation study to examine our proposed AutoTimes in Table~\ref{tab:ablation}. The results underscore that our method maintains the consistency of the decoder-only architecture and autoregressive inference, effectively leveraging LLMs and addressing the concerns regarding performance improvement and adaptation cost. Further, we provide a comparison by substituting our token-wise segment projection (consistent with LLMs) with the flatten linear head~\cite{nie2022time} (common in non-autoregressive forecasters). Results of Table~\ref{tab:ablation_supp} in the Appendix reveal that the performance of non-autoregressive generation is consistently inferior to that of our autoregressive AutoTimes approach. 

\begin{table}[htbp]
  \caption{We follow the protocol of LLM4TS ablation studies~\cite{tan2024language} to verify whether the LLM is truly useful in our AutoTimes: (1) \emph{w/o LLM} replaces the language model entirely and passing input tokens directly to the last layer; (2) \emph{LLM2Attn} replaces the language model with a single multi-head attention layer; (3) \emph{LLM2Trsf} replaces the language model with a single transformer block.}  \label{tab:ablation}
  \centering
  \begin{threeparttable}
  \begin{small}
  \renewcommand{\multirowsetup}{\centering}
  \setlength{\tabcolsep}{2pt}
  \resizebox{\textwidth}{!}{\begin{tabular}{l|c|cccccccc|cccccccc}
    \toprule
    \multicolumn{2}{c|}{Dataset} &  \multicolumn{8}{c|}{ETTh1}  &  \multicolumn{8}{c}{ECL}\\
    \cmidrule(lr){1-2}\cmidrule(lr){3-10} \cmidrule(lr){11-18} 
    \multicolumn{2}{c|}{Type} & \multicolumn{2}{c}{\scalebox{0.8}{\textbf{AutoTimes}}} &
    \multicolumn{2}{c}{\scalebox{0.80}{w/o LLM}}  & 
    \multicolumn{2}{c}{\scalebox{0.80}{LLM2Attn}} &
    \multicolumn{2}{c|}{\scalebox{0.80}{LLM2Trsf}}  & 
    \multicolumn{2}{c}{\scalebox{0.80}{\textbf{AutoTimes} }} &
    \multicolumn{2}{c}{\scalebox{0.80}{w/o LLM}} & 
    \multicolumn{2}{c}{\scalebox{0.80}{LLM2Attn}} & 
 \multicolumn{2}{c}{\scalebox{0.80}{LLM2Trsf}}  \\
     \cmidrule(lr){1-2}\cmidrule(lr){3-4} \cmidrule(lr){5-6}\cmidrule(lr){7-8} \cmidrule(lr){9-10}\cmidrule(lr){11-12}\cmidrule(lr){13-14}\cmidrule(lr){15-16}\cmidrule(lr){17-18}
    \multicolumn{2}{c|}{Metric} & \scalebox{0.90}{MSE} & \scalebox{0.90}{MAE} & \scalebox{0.90}{MSE} & \scalebox{0.90}{MAE} & \scalebox{0.90}{MSE} & \scalebox{0.90}{MAE} & \scalebox{0.90}{MSE} & \scalebox{0.90}{MAE} & \scalebox{0.90}{MSE} & \scalebox{0.90}{MAE} & \scalebox{0.90}{MSE} & \scalebox{0.90}{MAE} & \scalebox{0.90}{MSE} & \scalebox{0.90}{MAE} & \scalebox{0.90}{MSE} & \scalebox{0.90}{MAE}  \\
    \toprule
    \multicolumn{2}{c|}{\scalebox{0.90}{Pred-$96$}}
    & \scalebox{0.90}{\textbf{0.360}} & \scalebox{0.90}{\textbf{0.400}} 
    & \scalebox{0.90}{0.365} & \scalebox{0.90}{0.399} 
    & \scalebox{0.90}{0.383} & \scalebox{0.90}{0.404}    
    & \scalebox{0.90}{0.377} & \scalebox{0.90}{0.401} 
    & \scalebox{0.90}{\textbf{0.129}} & \scalebox{0.90}{\textbf{0.225}}
    & \scalebox{0.90}{0.171} & \scalebox{0.90}{0.263} 
    & \scalebox{0.90}{0.156} & \scalebox{0.90}{0.255} 
    & \scalebox{0.90}{0.162} & \scalebox{0.90}{0.263} \\
    
    \midrule
    \multicolumn{2}{c|}{\scalebox{0.90}{Pred-$192$}} 
    & \scalebox{0.90}{\textbf{0.388}} & \scalebox{0.90}{\textbf{0.419}} 
    & \scalebox{0.90}{0.405} & \scalebox{0.90}{0.425}
    & \scalebox{0.90}{0.414} & \scalebox{0.90}{0.422} 
    & \scalebox{0.90}{0.406} & \scalebox{0.90}{0.420}
    & \scalebox{0.90}{\textbf{0.147}} & \scalebox{0.90}{\textbf{0.241}}
    & \scalebox{0.90}{0.192} & \scalebox{0.90}{0.282} 
    & \scalebox{0.90}{0.178} & \scalebox{0.90}{0.276} 
    & \scalebox{0.90}{0.189} & \scalebox{0.90}{0.287} \\

    \midrule
    \multicolumn{2}{c|}{\scalebox{0.90}{Pred-$336$}} 
    & \scalebox{0.90}{\textbf{0.401}} & \scalebox{0.90}{\textbf{0.429}} 
    & \scalebox{0.90}{0.429} & \scalebox{0.90}{0.441} 
    & \scalebox{0.90}{0.431} & \scalebox{0.90}{0.432} 
    & \scalebox{0.90}{0.421} & \scalebox{0.90}{0.431}
    & \scalebox{0.90}{\textbf{0.162}} & \scalebox{0.90}{\textbf{0.258}} 
    & \scalebox{0.90}{0.216} & \scalebox{0.90}{0.304} 
    & \scalebox{0.90}{0.198} & \scalebox{0.9}{0.295} 
    & \scalebox{0.90}{0.216} & \scalebox{0.90}{0.309} \\
    
    \midrule
    \multicolumn{2}{c|}{\scalebox{0.90}{Pred-$720$}} 
    & \scalebox{0.90}{\textbf{0.406}} & \scalebox{0.90}{\textbf{0.440}} 
    & \scalebox{0.90}{0.450} & \scalebox{0.90}{0.468} 
    & \scalebox{0.90}{0.456} & \scalebox{0.90}{0.454} 
    & \scalebox{0.90}{0.449} & \scalebox{0.90}{0.452}
    & \scalebox{0.90}{\textbf{0.199}} & \scalebox{0.90}{\textbf{0.288}}
    & \scalebox{0.90}{0.264} & \scalebox{0.90}{0.342} 
    & \scalebox{0.90}{0.230} & \scalebox{0.90}{0.320} 
    & \scalebox{0.90}{0.258} & \scalebox{0.90}{0.340} \\
    \bottomrule
  \end{tabular}}
  \end{small}
  \end{threeparttable}
  \vspace{-5pt}
\end{table}

\paragraph{LoRA adaptation} By incorporating low-rank adaptation technique~\cite{hu2021lora} on the intermediate LLM layers, the token transition of the large language model can be further fine-tuned to align the future extrapolation of time series. Table~\ref{tab:gpt2_lora} provides the performance comparing the incorporation of LoRA, which consistently improves the performance of the LLM-based forecaster adapted by AutoTimes. 

\begin{table}[htbp]
  \caption{Full long-term forecasting results of AutoTimes and AutoTimes equipped with LoRA~\cite{hu2021lora}.}\label{tab:gpt2_lora}
  \centering
  \begin{threeparttable}
  \begin{small}
  \renewcommand{\multirowsetup}{\centering}
  \resizebox{\textwidth}{!}{\begin{tabular}{c|c|cc|cc|cc|cc|cc}
    \toprule
    \multicolumn{2}{c|}{Dataset} & 
    \multicolumn{2}{c}{\rotatebox{0}{\scalebox{1.0}{ETTh1}}} &
    \multicolumn{2}{c}{\rotatebox{0}{\scalebox{1.0}{ECL}}} &
    \multicolumn{2}{c}{\rotatebox{0}{\scalebox{1.0}{Weather}}} &
    \multicolumn{2}{c}{\rotatebox{0}{\scalebox{1.0}{Traffic}}} &
    \multicolumn{2}{c}{\rotatebox{0}{\scalebox{1.0}{Solar-Energy}}} \\
     \cmidrule(lr){1-2} \cmidrule(lr){3-4} \cmidrule(lr){5-6}\cmidrule(lr){7-8} \cmidrule(lr){9-10} \cmidrule(lr){11-12} 
    \multicolumn{2}{c|}{Metric} & \scalebox{1.0}{MSE} & \scalebox{1.0}{MAE}  & \scalebox{1.0}{MSE} & \scalebox{1.0}{MAE}  & \scalebox{1.0}{MSE} & \scalebox{1.0}{MAE}  & \scalebox{1.0}{MSE} & \scalebox{1.0}{MAE} & \scalebox{1.0}{MSE} & \scalebox{1.0}{MAE} \\
    \toprule
    \multirow{2}{*}{Pred-$96$} 
    & AutoTimes & 0.360 & \textbf{0.397} & 0.140 & 0.236 & 0.158 & 0.208 & 0.369 & 0.257 & 0.179 & 0.220 \\
    & \scalebox{1.0}{\textbf{+ LoRA}} & \textbf{0.357} & \textbf{0.397} & \textbf{0.130} & \textbf{0.225} & \textbf{0.151} & \textbf{0.201} & \textbf{0.360} & \textbf{0.256} & \textbf{0.176} & \textbf{0.219} \\
    \midrule
    \multirow{2}{*}{Pred-$192$} 
    & AutoTimes & \textbf{0.391} & \textbf{0.419} & 0.159 & 0.253 & 0.207 & 0.254 & 0.394 & 0.268 & 0.198 & 0.236 \\
    & \scalebox{1.0}{\textbf{+ LoRA}} & \textbf{0.391} & 0.420 & \textbf{0.149} & \textbf{0.242} & \textbf{0.197} & \textbf{0.244} & \textbf{0.383} & \textbf{0.267} & \textbf{0.195} & \textbf{0.235} \\
    \midrule
    \multirow{2}{*}{Pred-$336$} 
    & AutoTimes & \textbf{0.408} & \textbf{0.432} & 0.177 & 0.270 & 0.262 & 0.298 & 0.413 & 0.278 & 0.213 & 0.252 \\
    & \scalebox{1.0}{\textbf{+ LoRA}} & 0.409 & 0.433 & \textbf{0.164} & \textbf{0.259} & \textbf{0.251} & \textbf{0.287} & \textbf{0.401} & \textbf{0.277} & \textbf{0.208} & \textbf{0.249} \\
    \midrule
    \multirow{2}{*}{Pred-$720$} 
    & AutoTimes & 0.429 & 0.452 & 0.216 & 0.303 & 0.342 & 0.353 & 0.449 & \textbf{0.299} & 0.239 & 0.277 \\
    & \scalebox{1.0}{\textbf{+ LoRA}} & \textbf{0.426} & \textbf{0.451} & \textbf{0.202} & \textbf{0.293} & \textbf{0.326} & \textbf{0.339} & \textbf{0.440} & 0.300 & \textbf{0.225} & \textbf{0.268} \\
    \bottomrule
  \end{tabular}}
    \end{small}
  \end{threeparttable}
\vspace{-10pt}
\end{table}

\section{Conclusion}\label{sec:conclusion}
This paper aims to develop foundation models for time series forecasting. We utilize off-the-shelf LLMs as autoregressive forecasters by transferring the general-purpose and multi-step generation ability. Different from prior methods, we notice prevalent non-autoregressive LLM4TS methods may contradict the decoder-only structure and lead to insufficient utilization of LLMs. Experimentally, the proposed method achieves state-of-the-art performance with remarkable model efficiency. Further analysis reveals that our forecaster effectively inherits advanced capabilities such as zero-shot and in-context forecasting, and is able to utilize both instructive times series and timestamps. In the future, we will further incorporate advanced low-rank adaptation and utilize booming language backbones.

\newpage

\section*{Acknowledgments}
This work was supported by the Ministry of Industry and Information Technology of China.

\small
\bibliographystyle{plain}
\bibliography{ref}

\newpage
\appendix
\section{Dataset Descriptions}\label{sec:datasets}
We conduct experiments to evaluate the performance of the proposed AutoTimes on seven real-world datasets spanning diverse domains: (1) ETTh1~\cite{zhou2021informer} spans from July 2016 to July 2018 and consists of seven factors related to electricity transformers. (2) Weather ~\cite{wu2021autoformer} encompasses 21 meteorological factors collected every 10 minutes in 2020 from the Weather Station of the Max Planck Biogeochemistry Institute. (3) ECL~\cite{wu2021autoformer} captures hourly electricity consumption data from 321 clients. (4) Traffic ~\cite{wu2021autoformer} gathers hourly road occupancy rates from 862 sensors on San Francisco Bay area freeways, covering the period from January 2015 to December 2016. (5) Solar-Energy~\cite{lai2018modeling} records solar power production from 137 PV plants in 2006, sampled every 10 minutes. (6) M4 is a competition dataset encompassing various time series across different frequencies and domains such as business and economics. (7) M3, albeit smaller than M4, also contains diverse time series from various domains.

 We follow the same data processing and train-validation-test set split protocol used in TimesNet~\cite{wu2021autoformer}, where the train, validation, and test datasets are strictly divided according to chronological order to ensure no data leakage. As for long-term forecasting settings, we fix the context length of AutoTimes and the lookback length of other compared methods as 672 in ETT, ECL, Traffic, Weather, and Solar-Energy, and the forecast length varies in \{96, 192, 336, 720\}. For the short-term forecasting on M4 and M3 datasets, the input length is generally set to twice the output length according to the official implementation of TimesNet. The details are provided in Table~\ref{tab:dataset}.

 \begin{table}[htbp]

  \caption{Detailed dataset descriptions. \emph{Dim} denotes the variate number. \emph{Dataset Size} denotes the total number of time points in (Train, Validation, Test) splits respectively. \emph{Forecast Length} denotes the future time points to be predicted. \emph{Frequency} denotes the sampling interval of time points.}\label{tab:dataset}
  \vspace{-5pt}
  \vskip 0.05in
  \footnotesize
  \begin{threeparttable}
  \begin{small}
  \renewcommand{\multirowsetup}{\centering}
  \renewcommand\arraystretch{1.4}
  \setlength{\tabcolsep}{10pt}
  \resizebox{\textwidth}{!}{\begin{tabular}{c|c|c|c|c|c}
    \toprule
    Dataset & Dim & Forecast Length & Dataset Size & Frequency& Information \\
    \toprule
     ETTh1 & 7 & \{96, 192, 336, 720\} & (8545, 2881, 2881) & Hourly & Electricity\\
     \midrule
    Weather & 21 & \{96, 192, 336, 720\} & (36792, 5271, 10540) & 10min & Weather\\
    \midrule
    ECL & 321 & \{96, 192, 336, 720\} & (18317, 2633, 5261) & Hourly & Electricity \\
    \midrule
    Traffic & 862 & \{96, 192, 336, 720\} & (12185, 1757, 3509) & Hourly & Transportation \\
    \midrule
    Solar-Energy & 137  & \{96, 192, 336, 720\} & (36601, 5161, 10417) & 10min & Energy \\
    \midrule
    M4-Yearly & 1 & {6} & (23000, 0, 23000) & Yearly & Demographic \\
    \midrule
    M4-Quarterly & 1 & {8} & (24000, 0, 24000) & Quarterly & Finance \\
    \midrule
    M4-Monthly & 1 & {18} & (48000, 0, 48000) & Monthly & Industry \\
    \midrule
    M4-Weekly & 1 & {13} & (359, 0, 359) & Weekly & Macro \\
    \midrule
    M4-Daily & 1 & {14} & (4227, 0, 4227) & Daily & Micro \\
    \midrule
    M4-Hourly & 1 & 48 & (414, 0, 414) & Hourly & Other \\
    \midrule
    M3-Yearly & 1 & {6} & (645, 0, 645) & Yearly & Demographic \\
    \midrule
    M3-Quarterly & 1 & {8} & (756, 0, 756) & Quarterly & Finance \\
    \midrule
    M3-Monthly & 1 & {18} & (1428, 0, 1428) & Monthly & Industry \\
    \midrule
    M3-Others & 1 & {8} & (174, 0, 174) & Weekly & Macro \\
    \bottomrule
    \end{tabular}}
    \vspace{-5pt}
    \end{small}
    \end{threeparttable}
\end{table}

\section{Implementation Details}\label{sec:implementation}
AutoTimes processes timestamps in textual form rather than numerical encoding, potentially enabling to handle other textual data such as news or logs. We utilize LLM to obtain embedding for the special token \verb|<EOS>| to capture embedding for the entire sentence. Pseudo-code for this process is depicted in Algorithm~\ref{alg:textual_embedding}. It is worth noting that in the context of multivariate time series forecasting, timestamps are shared across variates. Thus, timestamps can implicitly express relationships between variates even with channel independence. Further, assuming there are $C$ variates since the number of timestamps is $\frac{1}{C}$ of the total time point count, these embeddings can be efficiently pre-computed by large language models. 

After obtaining embedding for the timestamps, we repurpose LLM for time series forecasting using Algorithm~\ref{alg:repurpose}. At this stage, only the parameters of $\operatorname{SegmentEmbedding}$ and $\operatorname{SegmentProjection}$ are updated, while the parameters of LLMs remain entirely frozen. During inference, AutoTimes utilizes the last token generated as its prediction and then employs this output to create subsequent predictions autoregressively. This approach enables AutoTimes to predict sequences of variable lengths with just one model dynamically. Such capability is crucial in real-world application scenarios. The pseudo-code in Algorithm~\ref{alg:forecast}-\ref{alg:rolling_forecast} illustrates this process.

All the experiments are conducted using PyTorch~\cite{paszke2019pytorch} on NVIDIA A100 GPUs. We employ Adam~\cite{kingma2014adam} with an initial learning rate in \{$10^{-3}, 5 \times 10^{-4}, 10^{-4}$\} and MSE loss for model optimization. We adopt Channel Independence~\cite{nie2022time} for multivariate time series and utilize our position embeddings of timestamps to explicitly align them. The batch size is chosen from $\{256, 1024, 2048\}$, and we set the number of training epochs as $10$. As for $\operatorname{SegmentEmbedding}$ and $\operatorname{SegmentProjection}$, we implement them by either a linear layer or MLP. Results of deep forecaster are based on the benchmark provided by the TimesNet~\cite{wu2021autoformer} repository, which is fairly built on the same configurations provided by the original paper. LLM4TS methods~\cite{jin2023time, liu2023unitime, zhou2023one} are implemented by their official and open-source repository. Unless otherwise specified, we use LLaMA-7B~\cite{touvron2023llama} as the default base LLM. We also present the standard deviation of AutoTimes forecasting performance with three random seeds in Table~\ref{tab:std}, demonstrating that the performance of AutoTimes is stable.

\begin{algorithm*}[htbp]
  \setstretch{1.5}
   \caption{AutoTimes - Generate Text Embedding}
   \label{alg:textual_embedding}
    \begin{algorithmic}[1]
       \Require  
       Input time series $\mathbf{x}_{(i-1)S+1:iS}$ \Comment $i$-th token of length $S$
       \State $\mathbf{s}_i = \{x_{(i-1)S+1},\dots,x_{iS}\}$  \Comment Model dimension of the LLM $D$
       \State $\mathbf{TE}_i = \operatorname{SelectLast}\big(\operatorname{LLM}(\operatorname{TextualDuration}(\mathbf{s}_i))\big)$ \Comment $\mathbf{TE}_{i} \in \mathbb{R}^{D}$
       \State $\textbf{Return}\ \mathbf{TE}_{i}$ \Comment Return textual embedding of $i$-th token
    \end{algorithmic}
\end{algorithm*}

\begin{algorithm}[htbp]
  \setstretch{1.5}
   \caption{AutoTimes - Repurpose LLM}
   \label{alg:repurpose}
\begin{algorithmic}[1]
   \Require Input time series $\{x_1, \dots, x_{(N+1)\times S}\}$; text embeddings $\{\mathbf{TE}_1,\dots, \mathbf{TE}_N\}$ \Comment Token number $N$
   \State $\textbf{for}\ i\ \textbf{in}\ \{1, \dots, N\}\textbf{:}$
   \State $\textbf{\textcolor{white}{for}}\ 
 \mathbf{s}_i = \{x_{(i-1)S+1},\dots,x_{iS}\}$ \Comment $\mathbf{s}_i \in \mathbb{R}^{S}$
   \State  $\textbf{\textcolor{white}{for}}\ 
 \mathbf{SE}_i = \operatorname{SegmentEmbedding}(\mathbf{s}_i)$ \Comment $\mathbf{SE}_i \in \mathbb{R}^{D}$
   \State  $\textbf{\textcolor{white}{for}}\ \mathbf{E}_i = \mathbf{SE}_i + \mathbf{TE}_i$ \Comment $\mathbf{E}_i \in \mathbb{R}^{D}$
   \State $\{\hat{\mathbf{E}}_2, \dots, \hat{\mathbf{E}}_{N+1}\} = \operatorname{LLMLayers}(\{\mathbf{E}_1, \dots, \mathbf{E}_N\})$ \Comment $\hat{\mathbf{E}}_i \in \mathbb{R}^{D}$
   \State $\textbf{for}\ i\ \textbf{in}\ \{2, \dots, N + 1\}\textbf{:}$
   \State $\textbf{\textcolor{white}{for}}\ \hat{\mathbf{s}}_i = \operatorname{SegmentProjection}(\hat{\mathbf{E}}_i)$ \Comment $\hat{\mathbf{s}}_i \in \mathbb{R}^{S}$
   \State $\mathbf{s}_{N+1} = \{x_{NS+1},\dots,x_{NS+S}\}$
   \State $\mathcal{L}_{\text{MSE}} = \frac{1}{NS} \sum ||\mathbf{s}_i-\hat{\mathbf{s}}_i||_2^2,\ i\in\{2,\dots,N+1\}$  \Comment Objective of the next token prediction 
\end{algorithmic}
\end{algorithm}

\begin{algorithm}[htbp]
  \setstretch{1.5}
   \caption{AutoTimes - LLM Forecasting}
   \label{alg:forecast}
\begin{algorithmic}[1]
   \Require Input time series $\{x_1, \dots, x_{N_1 \times S}\}$; text embeddings $\{\mathbf{TE}_1,\dots, \mathbf{TE}_{N_1}\}$ \Comment Lookback token number $N_1\le N$
   \State $\textbf{for}\ i\ \textbf{in}\ \{1, \dots, N_1\}\textbf{:}$
   \State $\textbf{\textcolor{white}{for}}\ 
 \mathbf{s}_i = \{x_{(i-1)S+1},\dots,x_{iS}\}$ \Comment $\mathbf{s}_i \in \mathbb{R}^{S}$
   \State  $\textbf{\textcolor{white}{for}}\ 
 \mathbf{SE}_i = \operatorname{SegmentEmbedding}(\mathbf{s}_i)$ \Comment $\mathbf{SE}_i \in \mathbb{R}^{D}$
   \State  $\textbf{\textcolor{white}{for}}\ \mathbf{E}_i = \mathbf{SE}_i + \mathbf{TE}_i$ \Comment $\mathbf{E}_i \in \mathbb{R}^{D}$
   \State $\{\hat{\mathbf{E}}_2, \dots, \hat{\mathbf{E}}_{N_1+1}\} = \operatorname{LLMLayers}(\{\mathbf{E}_1, \dots, \mathbf{E}_{N_1}\})$ \Comment $\hat{\mathbf{E}}_i \in \mathbb{R}^{D}$
   \State $\textbf{for}\ i\ \textbf{in}\ \{2, \dots, N + 1\}\textbf{:}$
   \State $\textbf{\textcolor{white}{for}}\ \hat{\mathbf{s}}_i = \operatorname{SegmentProjection}(\hat{\mathbf{E}}_i)$ \Comment $\hat{\mathbf{s}}_i \in \mathbb{R}^{S}$
   \State $\textbf{Return}$ $\hat{\mathbf{s}}_{N+1}$
  \Comment Return last token $\hat{\mathbf{s}}_{N+1} \in \mathbb{R}^{S}$ as the prediction
\end{algorithmic}
\end{algorithm}

\begin{algorithm}[htbp]
  \setstretch{1.5}
   \caption{AutoTimes - Autoregressive Generation}
   \label{alg:rolling_forecast}
\begin{algorithmic}[1]
   \Require Input time series  $\{x_1, \dots, x_{N_1 \times S}\}$; textual embeddings $\{\mathbf{TE}_1,\dots, \mathbf{TE}_{N_2}\}$ \Comment Forecast token number $N_2-N_1$
   \State $\mathbf{x} = \{x_1, \dots, x_{N_1 \times S}\}$
   \State prediction = \{\}
   \State $\textbf{for}\ i\ \textbf{in}\ \{1,\dots,N_2 - N_1\}\textbf{:}$
   \State $\textbf{\textcolor{white}{for}}\ \hat{\mathbf{y}} = \operatorname{LLMForecaster}(\mathbf{x}, \mathbf{TE}_{:N_1+i-1})$ \Comment Details in Algorithm~\ref{alg:forecast}
   \State $\textbf{\textcolor{white}{for}}\ \mathbf{x} \leftarrow \{\mathbf{x}, \hat{\mathbf{y}}\}$ \Comment Concatenate for the input for next iteration
   \State $\textbf{\textcolor{white}{for}}\ \text{prediction} \leftarrow \{\text{prediction}, \hat{\mathbf{y}}\}$ \Comment Record prediction results
   \State $\textbf{Return}$ prediction \Comment Return result $\in \mathbb{R}^{(N_2-N_1)\times S}$
\end{algorithmic}
\end{algorithm}

\begin{table}[htbp]
  \caption{Performance and standard deviations of AutoTimes. Results come from three random seeds.}
  \label{tab:std}
  \centering
  \begin{threeparttable}
  \begin{small}
  \renewcommand{\multirowsetup}{\centering}
  \setlength{\tabcolsep}{7pt}
  \resizebox{\textwidth}{!}{\begin{tabular}{c|cc|cc|cc}
    \toprule
    Dataset & \multicolumn{2}{c}{ETTh1} & \multicolumn{2}{c}{ECL} & \multicolumn{2}{c}{Weather}   \\
    \cmidrule(lr){1-1}\cmidrule(lr){2-3} \cmidrule(lr){4-5}\cmidrule(lr){6-7}
    Horizon & MSE & MAE & MSE & MAE & MSE & MAE \\
    \toprule
    $96$ & 0.360\scalebox{0.9}{$\pm$0.002} & 0.400\scalebox{0.9}{$\pm$0.002} & 0.129\scalebox{0.9}{$\pm$0.001} & 0.225\scalebox{0.9}{$\pm$0.001} & 0.153\scalebox{0.9}{$\pm$0.000} & 0.203\scalebox{0.9}{$\pm$0.001} \\
    $192$ & 0.388\scalebox{0.9}{$\pm$0.002} & 0.419\scalebox{0.9}{$\pm$0.001} & 0.147\scalebox{0.9}{$\pm$0.002} & 0.241\scalebox{0.9}{$\pm$0.002} & 0.201\scalebox{0.9}{$\pm$0.001} & 0.250\scalebox{0.9}{$\pm$0.001} \\
    $336$ & 0.401\scalebox{0.9}{$\pm$0.003} & 0.429\scalebox{0.9}{$\pm$0.002} & 0.162\scalebox{0.9}{$\pm$0.002} & 0.258\scalebox{0.9}{$\pm$0.003} & 0.256\scalebox{0.9}{$\pm$0.002} & 0.293\scalebox{0.9}{$\pm$0.001} \\
    $720$ & 0.406\scalebox{0.9}{$\pm$0.004} & 0.440\scalebox{0.9}{$\pm$0.002} & 0.199\scalebox{0.9}{$\pm$0.004} & 0.288\scalebox{0.9}{$\pm$0.002} & 0.331\scalebox{0.9}{$\pm$0.002} & 0.345\scalebox{0.9}{$\pm$0.001} \\
    \midrule
    Dataset &  \multicolumn{2}{c}{Traffic} & \multicolumn{2}{c}{Solar-Energy}\\
    \cmidrule(lr){1-1}\cmidrule(lr){2-3} \cmidrule(lr){4-5}
    Horizon & MSE & MAE & MSE & MAE  \\
    $96$ &
    0.343\scalebox{0.9}{$\pm$0.001} & 0.248\scalebox{0.9}{$\pm$0.001} &
    0.171\scalebox{0.9}{$\pm$0.001} & 0.221\scalebox{0.9}{$\pm$0.001}\\
    $192$ &
    0.362\scalebox{0.9}{$\pm$0.001} & 0.257\scalebox{0.9}{$\pm$0.001} &
    0.190\scalebox{0.9}{$\pm$0.001} & 0.236\scalebox{0.9}{$\pm$0.002} \\
    $336$ &
    0.379\scalebox{0.9}{$\pm$0.002} & 0.266\scalebox{0.9}{$\pm$0.001} &
    0.203\scalebox{0.9}{$\pm$0.002} & 0.248\scalebox{0.9}{$\pm$0.002} \\
    $720$ &
    0.413\scalebox{0.9}{$\pm$0.003} & 0.284\scalebox{0.9}{$\pm$0.002} &
    0.222\scalebox{0.9}{$\pm$0.002} & 0.262\scalebox{0.9}{$\pm$0.003} \\
    \bottomrule
  \end{tabular}}
  \end{small}
  \end{threeparttable}
\end{table}

\section{Hyperparameter Sensitivity}\label{sec:hyperparams}
$\operatorname{SegmentEmbedding}$ and $\operatorname{SegmentProjection}$ are uniformly implemented by MLP. The number of layers is fixed as $2$ and the hidden dimension is selected from $\{256, 512, 1024\}$ according to the validation loss. The segment length is set as $S=96$ in multivariate datasets and is set as the prediction length $S=F$ in M3 and M4. We verify the robustness of AutoTimes of hyperparameters as follows: the layer number and hidden dimension of $\operatorname{SegmentEmbedding}$ and $\operatorname{SegmentProjection}$, context length, and segment length. We observe that AutoTimes is insensitive to the configurations of embedding and projection layers. Besides, performance can be improved by increasing the context length. For long prediction lengths, a larger segment length is favored.

\begin{figure*}[h]
\begin{center}
    \center{\includegraphics[width=\textwidth]{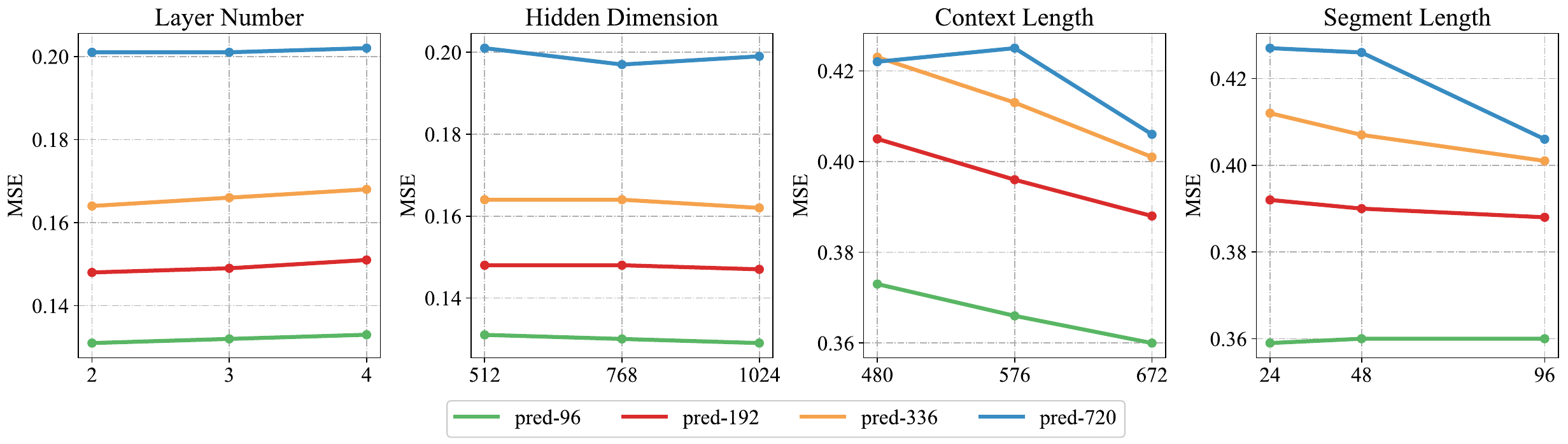}}
	\vspace{-15pt}
    \caption{Hyperparameter sensitivity of AutoTimes. Each curve presents a specific forecast length.}
	\label{fig:hyperparameter}
\end{center}
\vspace{-5pt}
\end{figure*}

\section{Supplementary Results}\label{sec:all_results}

\subsection{Time Series Forecasting}
We compare the performance of AutoTimes with state-of-the-art LLM4TS methods and well-acknowledged deep forecasters. Table~\ref{tab:forecast_short_full} shows detailed short-term forecast results on M4. Table~\ref{tab:one_model_full} presents results of the one-for-all forecasting benchmark across ETTh1, ECL, Traffic, Weather, and Solar-Energy datasets. We evaluate all methods by rolling forecasting: each model is trained with input length $672$ and output length $96$, and the predicted values are integrated as part of the input in the next iteration until reaching the desired forecast horizon.

Furthermore, the traditional one-for-one approach, where forecasters are trained individually for each prediction length, is also presented in Table~\ref{tab:one_for_one}. The results are reproduced using their corresponding official code. For the sake of rigor, we also provide our reproduced results with the officially reported results in Table~\ref{tab:reproduce}.

Additionally, we evaluate AutoTimes along with other baseline models on recent benchmarks~\cite{liu2024timer}. Results are presented in Table~\ref{tab:new_benchmark_results}. We also look forward to evaluating on more diverse benchmarks in the future.

\begin{table*}[htbp]
  \caption{Full long-term forecasting results of one-for-all: we conduct rolling forecasting with a single model trained on each dataset and accomplish four desired forecast lengths in $\{96, 192, 336, 720\}$. AutoTimes adapt LLMs with the context length $C=672$. We set the input length $L=672$ and output length $F=96$ in other methods, which are all implemented by their official code.}
  \label{tab:one_model_full}
  \vspace{-5pt}
  \vskip 0.15in
  \centering
  \begin{threeparttable}
  \begin{small}
  \renewcommand{\multirowsetup}{\centering}
    \setlength{\tabcolsep}{1pt}
  \renewcommand\arraystretch{1.4}
  \resizebox{\textwidth}{!}{\begin{tabular}{c|c|cc|cc|cc|cc|cc|cc|cc|cc}
    \toprule
    \multicolumn{2}{c}{Method}
    & \multicolumn{2}{c}{{\textbf{AutoTimes}}} & \multicolumn{2}{c}{\scalebox{0.9}{TimeLLM~\cite{jin2023time}}} &\multicolumn{2}{c}{\scalebox{0.9}{UniTime~\cite{liu2023unitime}}} &\multicolumn{2}{c}{FPT~\cite{zhou2023one}} & \multicolumn{2}{c}{iTrans.~\cite{liu2023itransformer}} & \multicolumn{2}{c}{DLinear~\cite{zeng2023transformers}} & \multicolumn{2}{c}{\scalebox{0.9}{PatchTST~\cite{nie2022time}}} & \multicolumn{2}{c}{\scalebox{0.9}{TimesNet~\cite{wu2022timesnet}}} \\ 
    \cmidrule(lr){1-2} \cmidrule(lr){3-4} \cmidrule(lr){5-6}\cmidrule(lr){7-8} \cmidrule(lr){9-10} \cmidrule(lr){11-12} \cmidrule(lr){13-14} \cmidrule(lr){15-16}\cmidrule(lr){17-18}
    \multicolumn{2}{c|}{Metric} & MSE & MAE & MSE & MAE & MSE & MAE & MSE & MAE & MSE & MAE  & MSE & MAE  & MSE & MAE  & MSE & MAE\\
    \toprule
    \multirow{5}{*}{\rotatebox{90}{ETTh1}}
    &  96 &  \textbf{0.360} & \textbf{0.400} & 0.380 & 0.412 & 0.386 & 0.409 & 0.377 & 0.404 & 0.387 & 0.418 & \underline{0.369} & \textbf{0.400} & 0.374 & \underline{0.401} & 0.452 & 0.463\\
    & 192 &  \textbf{0.388} & \textbf{0.419} & 0.408 & 0.431 & 0.639 & 0.577 & 0.412 & 0.426 & 0.416 & 0.437 & \underline{0.405} & \underline{0.422} & \underline{0.405} & \underline{0.422} & 0.474 & 0.477\\
    & 336 &  \textbf{0.401} & \textbf{0.429} & 0.425 & 0.443 & 0.814 & 0.680 & 0.440 & 0.444 & 0.434 & 0.450 & 0.435 & 0.445 & \underline{0.423} & \underline{0.435} & 0.493 & 0.489\\
    & 720 &  \textbf{0.406} & \textbf{0.440} & \underline{0.434} & 0.463 & 0.891 & 0.719 & 0.488 & 0.483 & 0.447 & 0.473 & 0.493 & 0.508 & \underline{0.434} & \underline{0.460} & 0.560 & 0.534\\
    \cmidrule(lr){2-18}
    & Avg &  \textbf{0.389} & \textbf{0.422} & 0.412 & 0.437 & 0.683 & 0.596 & 0.429 & 0.439 & 0.421 & 0.445 & 0.426 & 0.444 & \underline{0.409} & \underline{0.430} & 0.495 & 0.491\\
    \midrule
    \multirow{5}{*}{\rotatebox{90}{ECL}} 
    &  96  & \textbf{0.129} & \textbf{0.225} & 0.137 & 0.244 & 0.171 & 0.266 & 0.137 & 0.236 & 0.133 & \underline{0.229} & 0.138 & 0.238 & \underline{0.132} & 0.232 & 0.184 & 0.288\\
    &  192  & \textbf{0.147} & \textbf{0.241} & 0.158 & 0.266 & 0.293 & 0.378 & 0.158 & 0.258 & \underline{0.151} & \underline{0.245} & 0.152 & 0.251 & \underline{0.151} & 0.250 & 0.192 & 0.295 \\
    &  336  & \textbf{0.162} & \textbf{0.258} & 0.183 & 0.292 & 0.379 & 0.448 & 0.181 & 0.285 & 0.168 & \underline{0.262} & \underline{0.167} & 0.268 & 0.171 & 0.272 & 0.200 & 0.303\\
    &  720  & \textbf{0.199} & \textbf{0.288} & 0.247 & 0.348 & 0.455 & 0.502 & 0.258 & 0.355  & 0.205 & \underline{0.294} & \underline{0.203} & 0.302 & 0.222 & 0.318 & 0.228 & 0.325 \\
    \cmidrule(lr){2-18}
    & Avg  & \textbf{0.159} & \textbf{0.253} & 0.181 & 0.288 & 0.325 & 0.399 & 0.184 & 0.284 & \underline{0.164} & \underline{0.258} & 0.165 & 0.265 & 0.169 & 0.268 & 0.201 & 0.303\\
    \midrule
    \multirow{5}{*}{\rotatebox{90}{Weather}} 
    &  96 & \underline{0.153} & 0.203 & \textbf{0.149} & \textbf{0.200} & 0.180 & 0.223 & 0.154 & 0.205  & 0.174 & 0.225 & 0.169 & 0.229 & \textbf{0.149} & \underline{0.202} & 0.169 & 0.228\\
    &  192 & 0.201 & 0.250 & \textbf{0.193} & \textbf{0.243} & 0.450 & 0.451 & 0.196 & 0.243 & 0.227 & 0.268 & 0.211 & 0.268 & \underline{0.194} & \underline{0.245} & 0.222 & 0.269\\
    &  336& 0.256 & 0.293 & \textbf{0.243} & \textbf{0.284} & 0.594 & 0.570 & 0.244 & 0.282 & 0.290 & 0.309 & 0.258 & 0.306 & \underline{0.244} & \underline{0.285} & 0.290 & 0.310\\
    &  720 & 0.331 & 0.345 & \textbf{0.315} & \textbf{0.336} & 0.618 & 0.593 & 0.318 & 0.335 & 0.374 & 0.360 & 0.320 & 0.362 & \underline{0.317} & \underline{0.338} & 0.376 & 0.364\\
    \cmidrule(lr){2-18}
    & Avg & 0.235 & 0.273 & \textbf{0.225} & \textbf{0.266} & 0.461 & 0.459 & 0.228 & 0.266 & 0.266 & 0.291  & 0.239 & 0.291 & \underline{0.226} & \underline{0.268}  & 0.264 & 0.293\\
    \midrule
    \multirow{5}{*}{\rotatebox{90}{Traffic}} 
    &  96& \textbf{0.343} & \textbf{0.248} & 0.376 & 0.280 & 0.438 & 0.291 & 0.395 & 0.283 & \underline{0.353} & 0.259 & 0.399 & 0.285 & 0.359 & \underline{0.255} & 0.593 & 0.315\\
    &  192  & \textbf{0.362} & \textbf{0.257} & 0.397 & 0.294 & 0.538 & 0.353 & 0.425 & 0.302 & \underline{0.373} & 0.267 & 0.409 & 0.290 & 0.377 & \underline{0.265} & 0.596 & 0.317\\
    &  336  & \textbf{0.379} & \textbf{0.266} & 0.420 & 0.311 & 0.621 & 0.389 & 0.463 & 0.328 &  \underline{0.386} & \underline{0.275} & 0.422 & 0.297 & 0.393 & 0.276 & 0.600 & 0.319\\
    &  720 & \textbf{0.413} & \textbf{0.284} & 0.448 & 0.326 & 0.737 & 0.435 & 0.560 & 0.392  & \underline{0.425} & \underline{0.296} & 0.461 & 0.319 & 0.436 & 0.305 & 0.619 & 0.335\\
    \cmidrule(lr){2-18}
    & Avg & \textbf{0.374} & \textbf{0.264} & 0.410 & 0.303 & 0.584 & 0.367 & 0.461 & 0.326 & \underline{0.384} & \underline{0.274} & 0.423 & 0.298 & 0.391 & 0.275 & 0.602 & 0.322\\
    \midrule
    \multirow{5}{*}{\rotatebox{90}{Solar.}}  
    &  96  & \underline{0.171} & \textbf{0.221} & 0.224 & 0.289 & 0.223 & 0.274 & 0.196 & 0.261 & 0.183 & 0.265  & 0.193 & 0.258 & \textbf{0.168} & \underline{0.237} & 0.180 & 0.272\\
    &  192  & \underline{0.190} & \textbf{0.236} & 0.248 & 0.315 & 0.373 & 0.431 & 0.219 & 0.284  & 0.205 & 0.283  & 0.214 & 0.274 & \textbf{0.189} & \underline{0.257} & 0.199 & 0.286 \\
    &  336 & \textbf{0.203} & \textbf{0.248} & 0.269 & 0.338 & 0.445 & 0.536 & 0.245 & 0.311 & 0.224 & 0.299 & 0.233 & 0.291 &  \underline{0.212} & \underline{0.277} & 0.220 & 0.301\\
    &  720  & \textbf{0.222} & \textbf{0.262} & 0.310 & 0.396 & 0.526 & 0.608 & 0.285 & 0.354  & \underline{0.239} & 0.316 & 0.246 & 0.307 & 0.240 & \underline{0.305} & 0.251 & 0.321\\
    \cmidrule(lr){2-18}
    & Avg & \textbf{0.197} & \textbf{0.242} & 0.263 & 0.335 & 0.392 & 0.462 & 0.236 & 0.303 & 0.213 & 0.291 & 0.222 & 0.283 & \underline{0.202} & \underline{0.269} & 0.213 & 0.295\\
    \bottomrule
  \end{tabular}}
  \end{small}
  \end{threeparttable}
\end{table*}

\begin{table*}[htbp]
  \caption{Full results of short-term forecasting. We follow the same protocol of TimesNet~\cite{wu2022timesnet}.}
  \label{tab:forecast_short_full}
  \vspace{5pt}
  \centering
  \begin{threeparttable}
  \begin{small}
  \renewcommand{\multirowsetup}{\centering}
  \setlength{\tabcolsep}{3pt}
  \renewcommand\arraystretch{1.2}
  \begin{tabular}{c|c|cccccccccc}
    \toprule
    \multicolumn{2}{c|}{Method}   & 
    \multicolumn{1}{c}{\rotatebox{0}{\scalebox{0.85}{\textbf{AutoTimes}}}} &
    \multicolumn{1}{c}{\rotatebox{0}{\scalebox{0.85}{TimeLLM}}} &
    \multicolumn{1}{c}{\rotatebox{0}{\scalebox{0.85}{FPT}}} &
    \multicolumn{1}{c}{\rotatebox{0}{\scalebox{0.85}{Koopa}}} &
    \multicolumn{1}{c}{\rotatebox{0}{\scalebox{0.85}{N-HiTS}}} &
    \multicolumn{1}{c}{\rotatebox{0}{\scalebox{0.85}{{N-BEATS}}}} &
    \multicolumn{1}{c}{\rotatebox{0}{\scalebox{0.85}{PatchTST}}} &
    \multicolumn{1}{c}{\rotatebox{0}{\scalebox{0.85}{TimesNet}}} &
    \multicolumn{1}{c}{\rotatebox{0}{\scalebox{0.85}{DLinear}}} &
    \multicolumn{1}{c}{\rotatebox{0}{\scalebox{0.85}{FiLM}}} \\
    \toprule
    \multirow{3}{*}{\rotatebox{90}{Yearly}}
    & SMAPE & \textbf{13.319} & 13.419 & 13.531 & \underline{13.352}  & 13.371 & 13.866 & 13.517 & 13.394 & 14.012 & 13.466\\
    & MASE  & \textbf{2.993} & 3.005 & 3.015 & \underline{2.997}  & 3.025 & 3.006 & 3.031 & 3.004 & 3.071 & 3.059\\
    & OWA   & \textbf{0.784} & 0.789 & 0.793 & \underline{0.786} & 0.790 & 0.802 & 0.795 & 0.787 & 0.815 & 0.797\\
    \midrule
    \multirow{3}{*}{\rotatebox{90}{Quarterly}}
    & SMAPE & \underline{10.101} & 10.110 & 10.177 & 10.159 & 10.454 & 10.689 & 10.847  & \underline{10.101} & 10.758 & \textbf{10.074}\\
    & MASE  & 1.182 & \underline{1.178} & 1.194 & 1.189  & 1.219 & 1.294 & 1.315 & 1.183 & 1.306 & \textbf{1.163}\\
    & OWA  & 0.890 & \underline{0.889} & 0.897 & 0.895 & 0.919 & 0.957 & 0.972  & 0.890 & 0.905 & \textbf{0.881}\\
    \midrule
    \multirow{3}{*}{\rotatebox{90}{Monthly}}
    & SMAPE & \textbf{12.710} & 12.980 & 12.894 & \underline{12.730}  & 12.794 & 13.372 & 14.584 & 12.866 & 13.377 & 12.801\\
    & MASE & \textbf{0.934} & 0.963 &  0.956 & \underline{0.953}  & 0.960 & 1.014 & 1.169  & 0.964 & 1.021 & 0.955\\
    & OWA  & \textbf{0.880} & 0.903 & 0.897 & 0.901 & 0.895 & 0.940 & 1.055 & 0.894 & 0.944 & \underline{0.893}\\
    \midrule
    \multirow{3}{*}{\rotatebox{90}{Others}}
    & SMAPE & 4.843 & \underline{4.795} & 4.940 & 4.861 & \textbf{4.696} & 4.894 & 6.184  & 4.982 & 5.259 & 5.008\\
    & MASE & 3.277 & 3.178 & 3.228 & \textbf{3.124} & \underline{3.130} & 3.358 & 4.818 & 3.323 & 3.608 & 3.443\\
    & OWA  & 1.026 & 1.006 & 1.029 & \underline{1.004} & \textbf{0.988} & 1.044 & 1.140 & 1.048 & 1.122 & 1.070\\
    \midrule
    \multirow{3}{*}{\rotatebox{90}{Average}}
    & SMAPE  & \textbf{11.831} & 11.983 & 11.991 & \underline{11.863} & 11.960 & 12.418 & 13.022 & 11.930 & 12.489 & 11.910\\
    & MASE & \textbf{1.585} & \underline{1.595} & 1.600 & \underline{1.595}  & 1.606 & 1.656 & 1.814 & 1.597 & 1.690 & 1.613\\
    & OWA  & \textbf{0.850} & 0.859 & 0.861 & \underline{0.858} & 0.861 & 0.891 & 0.954  & 0.867 & 0.902 & 0.862\\
    \bottomrule
  \end{tabular}
  \end{small}
  \end{threeparttable}
\end{table*}

\begin{table*}[htbp]
  \caption{Long-term forecasting results of one-for-one: AutoTimes trains one LLM-based forecaster to handle all prediction lengths by autoregression, whereas other models are trained respectively on each prediction length. AutoTimes adapts LLMs with the context length $C = 672$. The lookback length is set as $L = 672$ in others. All results are averaged. Full results is provided in Table~\ref{tab:one_for_one_full}.}
  \vspace{5pt}
  \label{tab:one_for_one}
  \centering
  \begin{threeparttable}
  \begin{small}
  \renewcommand{\multirowsetup}{\centering}
  \setlength{\tabcolsep}{2pt}
  \resizebox{\textwidth}{!}{\begin{tabular}{c|c|cccccccccccccccc}
    \toprule
    \multicolumn{2}{c|}{Models} & \multicolumn{2}{c}{\scalebox{0.9}{\textbf{AutoTimes} }} &
    \multicolumn{2}{c}{\scalebox{0.80}{TimeLLM~\cite{jin2023time}}}  & 
    \multicolumn{2}{c}{\scalebox{0.80}{UniTime~\cite{liu2023unitime}}} &
    \multicolumn{2}{c}{\scalebox{0.90}{FPT~\cite{zhou2023one}}}  & 
    \multicolumn{2}{c}{\scalebox{0.80}{iTrans.~\cite{liu2023itransformer}}} &
    \multicolumn{2}{c}{\scalebox{0.80}{DLinear~\cite{zeng2023transformers}}} & 
    \multicolumn{2}{c}{\scalebox{0.75}{PatchTST~\cite{nie2022time}}} & 
 \multicolumn{2}{c}{\scalebox{0.75}{TimesNet~\cite{wu2022timesnet}}}  \\
    \cmidrule(lr){1-2}\cmidrule(lr){3-4} \cmidrule(lr){5-6}\cmidrule(lr){7-8} \cmidrule(lr){9-10}\cmidrule(lr){11-12}\cmidrule(lr){13-14}\cmidrule(lr){15-16}\cmidrule(lr){17-18}
    \multicolumn{2}{c|}{Metric} & \scalebox{0.90}{MSE} & \scalebox{0.90}{MAE} & \scalebox{0.90}{MSE} & \scalebox{0.90}{MAE} & \scalebox{0.90}{MSE} & \scalebox{0.90}{MAE} & \scalebox{0.90}{MSE} & \scalebox{0.90}{MAE} & \scalebox{0.90}{MSE} & \scalebox{0.90}{MAE} & \scalebox{0.90}{MSE} & \scalebox{0.90}{MAE} & \scalebox{0.90}{MSE} & \scalebox{0.90}{MAE} & \scalebox{0.90}{MSE} & \scalebox{0.90}{MAE}  \\
    \toprule
    \multicolumn{2}{c|}{\scalebox{0.90}{ETTh1}}
    & \scalebox{0.90}{\textbf{0.389}} & \scalebox{0.90}{\textbf{0.422}} 
    & \scalebox{0.90}{\underline{0.409}} & \scalebox{0.90}{0.432} 
    & \scalebox{0.90}{0.438} & \scalebox{0.90}{0.445}    
    & \scalebox{0.90}{0.426} & \scalebox{0.90}{0.438} 
    & \scalebox{0.90}{0.438} & \scalebox{0.90}{0.450}
    & \scalebox{0.90}{0.423} & \scalebox{0.90}{0.437} 
    & \scalebox{0.90}{0.413} & \scalebox{0.90}{\underline{0.431}} 
    & \scalebox{0.90}{0.458} & \scalebox{0.90}{0.450} \\
    
    \midrule
    \multicolumn{2}{c|}{\scalebox{0.90}{ECL}} 
    & \scalebox{0.90}{\textbf{0.159}} & \scalebox{0.90}{\textbf{0.253}} 
    & \scalebox{0.90}{0.170} & \scalebox{0.90}{0.275}
    & \scalebox{0.90}{0.194} & \scalebox{0.90}{0.287} 
    & \scalebox{0.90}{0.167} & \scalebox{0.90}{0.264}
    & \scalebox{0.90}{\underline{0.161}} & \scalebox{0.90}{\underline{0.256}}
    & \scalebox{0.90}{0.177} & \scalebox{0.90}{0.274} 
    & \scalebox{0.90}{\textbf{0.159}} & \scalebox{0.90}{\textbf{0.253}} 
    & \scalebox{0.90}{0.192} & \scalebox{0.90}{0.295} \\

    \midrule
    \multicolumn{2}{c|}{\scalebox{0.90}{Weather}} 
    & \scalebox{0.90}{0.235} & \scalebox{0.90}{0.273} 
    & \scalebox{0.90}{\textbf{0.227}} & \scalebox{0.90}{\underline{0.266}} 
    & \scalebox{0.90}{0.260} & \scalebox{0.90}{0.283} 
    & \scalebox{0.90}{0.231} & \scalebox{0.90}{0.269}  
    & \scalebox{0.90}{0.238} & \scalebox{0.90}{0.272} 
    & \scalebox{0.90}{0.240} & \scalebox{0.90}{0.300} 
    & \scalebox{0.90}{\underline{0.226}} & \scalebox{0.90}{\textbf{0.264}} 
    & \scalebox{0.90}{0.259} & \scalebox{0.90}{0.287} \\
    
    \midrule
    \multicolumn{2}{c|}{\scalebox{0.90}{Traffic}} 
    & \scalebox{0.90}{\textbf{0.374}} & \scalebox{0.90}{\textbf{0.264}} 
    & \scalebox{0.90}{0.402} & \scalebox{0.90}{0.294} 
    & \scalebox{0.90}{0.460} & \scalebox{0.90}{0.301} 
    & \scalebox{0.90}{0.416} & \scalebox{0.90}{0.295}
    & \scalebox{0.90}{\underline{0.379}} & \scalebox{0.90}{\underline{0.272}}
    & \scalebox{0.90}{0.434} & \scalebox{0.90}{0.295} 
    & \scalebox{0.90}{0.391} & \scalebox{0.90}{\textbf{0.264}} 
    & \scalebox{0.90}{0.620} & \scalebox{0.90}{0.336} \\

    \midrule
    \multicolumn{2}{c|}{\scalebox{0.90}{Solar.}} 
    & \scalebox{0.90}{\underline{0.197}} & \scalebox{0.90}{\textbf{0.242}}  
    &  \scalebox{0.90}{0.234} & \scalebox{0.90}{0.293}
    & \scalebox{0.90}{0.254} & \scalebox{0.90}{0.291}
    & \scalebox{0.90}{0.229} & \scalebox{0.90}{0.296}
    & \scalebox{0.90}{0.202} & \scalebox{0.90}{0.269}  
    & \scalebox{0.90}{0.217} & \scalebox{0.90}{0.278} 
    & \scalebox{0.90}{\textbf{0.189}} & \scalebox{0.90}{\underline{0.257}} 
    & \scalebox{0.90}{0.200} & \scalebox{0.90}{0.268} \\
    \bottomrule
  \end{tabular}}
  \end{small}
  \end{threeparttable}
\end{table*}

\begin{table}[htbp]
  \caption{Results of LLM4TS methods from the original paper and our reproduction by official code.}
  \vspace{-2pt}
  \label{tab:reproduce}
  \centering
  \begin{threeparttable}
  \begin{small}
  \renewcommand{\multirowsetup}{\centering}
  \setlength{\tabcolsep}{2pt}
  \resizebox{\textwidth}{!}{\begin{tabular}{c|c|cccccccccccccc}
    \toprule
    \multicolumn{2}{c|}{Models} & \multicolumn{2}{c}{\scalebox{0.9}{\textbf{AutoTimes} }} &
    \multicolumn{2}{c}{\scalebox{0.80}{TimeLLM$^*$~\cite{jin2023time}}}  & 
    \multicolumn{2}{c}{\scalebox{0.80}{TimeLLM~\cite{jin2023time}}} &
    \multicolumn{2}{c}{\scalebox{0.90}{FPT$^*$~\cite{zhou2023one}}}  & 
    \multicolumn{2}{c}{\scalebox{0.90}{FPT~\cite{zhou2023one}}}  & 
    \multicolumn{2}{c}{\scalebox{0.80}{UniTime$^*$~\cite{liu2023unitime}}} &
    \multicolumn{2}{c}{\scalebox{0.80}{UniTime~\cite{liu2023unitime}}} \\
    \cmidrule(lr){1-2}\cmidrule(lr){3-4} \cmidrule(lr){5-6}\cmidrule(lr){7-8} \cmidrule(lr){9-10}\cmidrule(lr){11-12}\cmidrule(lr){13-14}\cmidrule(lr){15-16}
    \multicolumn{2}{c|}{Metric} & \scalebox{0.90}{MSE} & \scalebox{0.90}{MAE} & \scalebox{0.90}{MSE} & \scalebox{0.90}{MAE} & \scalebox{0.90}{MSE} & \scalebox{0.90}{MAE} & \scalebox{0.90}{MSE} & \scalebox{0.90}{MAE} & \scalebox{0.90}{MSE} & \scalebox{0.90}{MAE} & \scalebox{0.90}{MSE} & \scalebox{0.90}{MAE} & \scalebox{0.90}{MSE} & \scalebox{0.90}{MAE} \\
    \toprule
    \multicolumn{2}{c|}{\scalebox{0.90}{ETTh1}}
    & \scalebox{0.90}{\textbf{0.389}} & \scalebox{0.90}{\textbf{0.422}} 
    & \scalebox{0.90}{\underline{0.408}} & \scalebox{0.90}{\underline{0.423}} 
    & \scalebox{0.90}{0.409} & \scalebox{0.90}{0.432} 
    & \scalebox{0.90}{0.427} & \scalebox{0.90}{0.426} 
    & \scalebox{0.90}{0.426} & \scalebox{0.90}{0.438}
    & \scalebox{0.90}{0.442} & \scalebox{0.90}{0.448}    
    & \scalebox{0.90}{0.438} & \scalebox{0.90}{0.445}    \\
    
    \midrule
    \multicolumn{2}{c|}{\scalebox{0.90}{ECL}} 
    & \scalebox{0.90}{\textbf{0.159}} & \scalebox{0.90}{\textbf{0.253}} 
    & \scalebox{0.90}{\textbf{0.159}} & \scalebox{0.90}{\textbf{0.253}}
    & \scalebox{0.90}{0.170} & \scalebox{0.90}{0.275}
    & \scalebox{0.90}{\underline{0.167}} & \scalebox{0.90}{\underline{0.263}}
    & \scalebox{0.90}{0.167} & \scalebox{0.90}{0.264}
    & \scalebox{0.90}{0.216} & \scalebox{0.90}{0.305} 
    & \scalebox{0.90}{0.194} & \scalebox{0.90}{0.287} \\

    \midrule
    \multicolumn{2}{c|}{\scalebox{0.90}{Weather}} 
    & \scalebox{0.90}{0.235} & \scalebox{0.90}{0.273} 
    & \scalebox{0.90}{\textbf{0.225}} & \scalebox{0.90}{\textbf{0.257}}
    & \scalebox{0.90}{\textbf{0.227}} & \scalebox{0.90}{\underline{0.266}} 
    & \scalebox{0.90}{0.237} & \scalebox{0.90}{0.270}  
    & \scalebox{0.90}{\underline{0.231}} & \scalebox{0.90}{\underline{0.269}} 
    & \scalebox{0.90}{0.253} & \scalebox{0.90}{0.276}
    & \scalebox{0.90}{0.260} & \scalebox{0.90}{0.283} \\
    
    \midrule
    \multicolumn{2}{c|}{\scalebox{0.90}{Traffic}} 
    & \scalebox{0.90}{\textbf{0.374}} & \scalebox{0.90}{\textbf{0.264}} 
    & \scalebox{0.90}{\underline{0.388}} & \scalebox{0.90}{\textbf{0.264}} 
    & \scalebox{0.90}{0.402} & \scalebox{0.90}{0.294} 
    & \scalebox{0.90}{0.414} & \scalebox{0.90}{\underline{0.294}}
    & \scalebox{0.90}{0.416} & \scalebox{0.90}{0.295}
    & \scalebox{0.90}{-} & \scalebox{0.90}{-} 
    & \scalebox{0.90}{0.460} & \scalebox{0.90}{0.301}  \\
    
    \midrule
    \multicolumn{2}{c|}{\scalebox{0.90}{Solar.}} 
    & \scalebox{0.90}{\textbf{0.197}} & \scalebox{0.90}{\textbf{0.242}}
    &  \scalebox{0.90}{-} & \scalebox{0.90}{-}
    &  \scalebox{0.90}{0.234} & \scalebox{0.90}{0.293}
    &  \scalebox{0.90}{-} & \scalebox{0.90}{-}
    & \scalebox{0.90}{\underline{0.229}} & \scalebox{0.90}{\underline{0.296}}
    &  \scalebox{0.90}{-} & \scalebox{0.90}{-}
    & \scalebox{0.90}{0.254} & \scalebox{0.90}{0.291} \\
    \bottomrule
  \end{tabular}}
  \end{small}
   \begin{tablenotes}
        \footnotesize
        \item[1] Methods with $^*$ means results from the original paper; without $^*$  means the reproduction.
        \item[2] ``-'' indicates that results are not reported in the original paper.
  \end{tablenotes}
  \end{threeparttable}
\end{table}

\begin{table}[htbp]
  \caption{Full long-term forecasting results of one-for-one: AutoTimes trains one LLM-based forecaster to handle all prediction lengths by autoregression, whereas other models are trained respectively on each prediction length. AutoTimes adapt LLMs with the context length $C = 672$. The lookback length is set as $L = 672$ in others, which are all implemented by their official code.}
  \label{tab:one_for_one_full}
  \centering
  \begin{threeparttable}
  \begin{small}
  \renewcommand{\multirowsetup}{\centering}
  \setlength{\tabcolsep}{1.2pt}
  \renewcommand\arraystretch{1.3}
  \resizebox{\textwidth}{!}{
  \begin{tabular}{c|c|cc|cc|cc|cc|cc|cc|cc|cc}
    \toprule
    \multicolumn{2}{c}{\multirow{2}{*}{Method}} & \multicolumn{2}{c}{One-for-all} & \multicolumn{14}{c}{Trained respectively on specific lookback/prediction length} \\
    \cmidrule(lr){3-4} \cmidrule(lr){5-18}
    \multicolumn{2}{c|}{} & \multicolumn{2}{c}{\scalebox{0.9}{\textbf{AutoTimes} }} &
    \multicolumn{2}{c}{\scalebox{0.80}{TimeLLM~\cite{jin2023time}}}  & 
    \multicolumn{2}{c}{\scalebox{0.80}{UniTime~\cite{liu2023unitime}}} &
    \multicolumn{2}{c}{\scalebox{0.80}{FPT~\cite{zhou2023one}}}  & 
    \multicolumn{2}{c}{\scalebox{0.80}{iTrans.~\cite{liu2023itransformer}}} &
    \multicolumn{2}{c}{\scalebox{0.80}{DLinear~\cite{zeng2023transformers}}} & 
    \multicolumn{2}{c}{\scalebox{0.80}{PatchTST~\cite{nie2022time}}} & 
    \multicolumn{2}{c}{\scalebox{0.80}{TimesNet~\cite{wu2022timesnet}}}  \\
    
    \cmidrule(lr){1-2} \cmidrule(lr){3-4} \cmidrule(lr){5-6}\cmidrule(lr){7-8} \cmidrule(lr){9-10} \cmidrule(lr){11-12} \cmidrule(lr){13-14}\cmidrule(lr){15-16} \cmidrule(lr){17-18}
    \multicolumn{2}{c|}{Metric} & MSE & MAE & MSE & MAE & MSE & MAE & MSE & MAE & MSE & MAE & MSE & MAE & MSE & MAE & MSE & MAE \\
    \toprule
    \multirow{5}{*}{\rotatebox{90}{ETTh1}}
    &  96 &  \textbf{0.360} & \underline{0.400} & 0.380 & 0.412 & 0.386 & 0.409 & 0.377 & 0.404 & 0.386 & 0.405 & 0.375 & \textbf{0.399} & \underline{0.370} & \textbf{0.399} & 0.384 & 0.402\\
    & 192 &  \textbf{0.388} & \underline{0.419} & \underline{0.405} & 0.422 & 0.428 & 0.436 & 0.413 & 0.424 &
     0.422 & 0.439 & \underline{0.405} & \textbf{0.416} & 0.413 & 0.421 & 0.557 & 0.436\\
    & 336 &  \textbf{0.401} & \textbf{0.429} & \underline{0.422} & \underline{0.433} & 0.464 & 0.456 & 0.436 & 0.444 & 0.444 & 0.457 & 0.439 & 0.443 & \underline{0.422} & 0.436 & 0.491 & 0.469\\
    & 720 &  \textbf{0.406} & \textbf{0.440} & \underline{0.430} & \underline{0.459} & 0.473 & 0.479 & 0.477 & 0.481 & 0.500 & 0.498 & 0.472 & 0.490 & 0.447 & 0.466 & 0.521 & 0.500\\
    \cmidrule(lr){2-18}
    & Avg &  \textbf{0.389} & \textbf{0.422} & \underline{0.409} & 0.432 & \underline{0.438} & 0.445 & 0.426 & 0.438 & 0.438 & 0.450 & 0.423 & 0.437 & 0.413 & \underline{0.431} & 0.458 & 0.450\\
    \midrule 
    \multirow{5}{*}{\rotatebox{90}{ECL}} 
    &  96  & \textbf{0.129} & \underline{0.225} & 0.137 & 0.244 & 0.171 & 0.266 & 0.137 & 0.236 & \underline{0.132} & 0.227 & 0.153 & 0.237 & \textbf{0.129} & \textbf{0.222} & 0.168 & 0.272\\
    &  192  & \textbf{0.147} & \underline{0.241} & 0.162 & 0.271 & 0.178 & 0.274 & 0.154 & 0.251 & 0.153 & 0.249 & 0.152 & 0.249 & \textbf{0.147} & \textbf{0.240} & 0.184 & 0.289\\
    &  336  & \textbf{0.162} & \textbf{0.258} & 0.175 & 0.279 & 0.194 & 0.289 & 0.169 & 0.267 & 0.167 & 0.262 & 0.169 & 0.267 & \underline{0.163} & \underline{0.259} & 0.198 & 0.300\\
    &  720  & 0.199 & \underline{0.288} & 0.207 & 0.306 & 0.232 & 0.319 & 0.207 & 0.300 & \textbf{0.192} & \textbf{0.285} & 0.233 & 0.344 & \underline{0.197} & 0.290 & 0.220 & 0.320\\
    \cmidrule(lr){2-18}
    & Avg  & \textbf{0.159} & \textbf{0.253} & 0.170 & 0.275 & 0.194 & 0.287 & 0.167 & 0.264 & \underline{0.161} & \underline{0.256} & 0.177 & 0.274 & \textbf{0.159} & \textbf{0.253} & 0.192 & 0.295\\
    \midrule
    \multirow{5}{*}{\rotatebox{90}{Weather}} 
    &  96 & 0.153 & 0.203 & \textbf{0.149} & \underline{0.200} & 0.180 & 0.223 & 0.154 & 0.205 & 0.163 & 0.211 & \underline{0.152} & 0.237 & \textbf{0.149} & \textbf{0.198} & 0.172 & 0.220\\
    &  192 & 0.201 & 0.250  & \underline{0.195} & \underline{0.243} & 0.226 & 0.261 & 0.196 & 0.245 & 0.205 & 0.250 & 0.220 & 0.282 & \textbf{0.194} & \textbf{0.241} & 0.219 & 0.261\\
    &  336& 0.256 & 0.293 & \textbf{0.245} & \textbf{0.282} & 0.280 & 0.300 & \underline{0.254} & 0.290 & \underline{0.254} & \underline{0.289} & 0.265 & 0.319 & \textbf{0.245} & \textbf{0.282} & 0.280 & 0.306\\
    &  720 & 0.331 & 0.345 & \underline{0.318} & 0.338 & 0.355 & 0.348 & 0.321 & \underline{0.337} & 0.329 & 0.340 & 0.323 & 0.362 & \textbf{0.314} & \textbf{0.334} & 0.365 & 0.359\\
    \cmidrule(lr){2-18}
    & Avg & 0.235 & 0.273 & \underline{0.227} & \underline{0.266} & 0.260 & 0.283 & 0.231 & 0.269 & 0.238 & 0.272 & 0.240 & 0.300 & \textbf{0.226} & \textbf{0.264} & 0.259 & 0.287\\
    \midrule
    \multirow{5}{*}{\rotatebox{90}{Traffic}} 
    &  96& \textbf{0.343} & \textbf{0.248} & 0.373 & 0.280 & 0.438 & 0.291 & 0.395 & 0.283 & \underline{0.351} & 0.257 & 0.410 & 0.282 & 0.360 & \underline{0.249} & 0.593 & 0.321\\
    &  192  & \textbf{0.362} & \underline{0.257} & 0.390 & 0.288 & 0.446 & 0.293 & 0.410 & 0.290 & \underline{0.364} & 0.265 & 0.423 & 0.287 & 0.379 & \textbf{0.256} & 0.617 & 0.336\\
    &  336  & \textbf{0.379} & \underline{0.266} & 0.407 & 0.299 & 0.461 & 0.300 & 0.414 & 0.295  & \underline{0.382} & 0.273 & 0.436 & 0.296 & 0.392 & \textbf{0.264} & 0.629 & 0.336\\
    &  720 & \textbf{0.413} & \textbf{0.284} & 0.438 & 0.310 & 0.494 & 0.318 & 0.445 & 0.311 & \underline{0.420} & 0.292 & 0.466 & 0.315 & 0.432 & \underline{0.286} & 0.640 & 0.350\\
    \cmidrule(lr){2-18}
    & Avg & \textbf{0.374} & \textbf{0.264} & 0.402 & 0.294 & 0.460 & 0.301 & 0.416 & 0.295 & \underline{0.379} & \underline{0.272} & 0.434 & 0.295 & 0.391 & \textbf{0.264} & 0.620 & 0.336\\
    \midrule
    \multirow{5}{*}{\rotatebox{90}{Solar-Energy}}  
    &  96  & \underline{0.171} & \textbf{0.221}   & 0.224 & 0.289 & 0.223 & 0.274  & 0.196 & 0.261  & 0.187 & 0.255  & 0.191 & 0.256 & \textbf{0.168} & \underline{0.237} & 0.178 & 0.256\\
    &  192 & \underline{0.190} & \textbf{0.236}   & 0.244 & 0.289 & 0.251 & 0.290  & 0.224 & 0.292  & 0.200 & 0.270  & 0.211 & 0.273 & \textbf{0.187} & \underline{0.263} & 0.200 & 0.268\\
    &  336 & \underline{0.203} & \textbf{0.248}  & 0.225 & 0.291 & 0.270 & 0.301  & 0.240 & 0.308  & 0.209 & 0.276  & 0.228 & 0.287 & \textbf{0.196} & \underline{0.260} & 0.212 & 0.274\\
    &  720 & 0.222 & \textbf{0.262} & 0.243 & 0.301  & 0.271 & 0.298  & 0.256 & 0.321  & 0.213 & 0.276  & 0.236 & 0.295 & \textbf{0.205} & \underline{0.269} & \underline{0.211} & 0.273\\
    \cmidrule(lr){2-18}
    & Avg & \underline{0.197} & \textbf{0.242} & 0.234 & 0.293 & 0.254 & 0.291  & 0.229 & 0.296  & 0.202 & 0.269  & 0.217 & 0.278 & \textbf{0.189} & \underline{0.257} & 0.200 & 0.268\\
    \bottomrule
  \end{tabular}}
  \end{small}
  \end{threeparttable}
\end{table}

\begin{table*}[htbp]
  \caption{Forecasting results on additional benchmark datasets~\cite{liu2024timer} ($672$-pred-$96$).}
  \vspace{5pt}
  \label{tab:new_benchmark_results}
  \centering
  \begin{threeparttable}
  \begin{small}
  \renewcommand{\multirowsetup}{\centering}
  \setlength{\tabcolsep}{8pt}
  \resizebox{\textwidth}{!}{\begin{tabular}{c|cc|cc|cc|cc}
    \toprule
    Models & \multicolumn{2}{c|}{\textbf{AutoTimes}} & \multicolumn{2}{c|}{PatchTST} & \multicolumn{2}{c|}{iTransformer} & \multicolumn{2}{c}{DLinear}  \\ 
    \cmidrule(lr){1-1} \cmidrule(lr){2-3} \cmidrule(lr){4-5} \cmidrule(lr){6-7} \cmidrule(lr){8-9}
    Metric & MSE & MAE & MSE & MAE & MSE & MAE & MSE & MAE   \\
    \toprule
    Australian Electricity &  \textbf{0.150} & \textbf{0.228} & 0.163 & 0.242 & 0.153 & 0.233 & 0.167 & 0.250    \\
    \midrule
    Bdg-2 Panther &  \textbf{0.537} & \textbf{0.458} & 0.565 & 0.476 & 0.546 & 0.462 & 0.581 & 0.499    \\
    \midrule
    Oikolab Weather &  \textbf{0.603} & \textbf{0.577} & 0.635 & 0.603 & 0.630 & 0.591 & 0.663 & 0.611    \\
    \bottomrule
  \end{tabular}}
  \end{small}
  \end{threeparttable}
\end{table*}

\subsection{Zero-Shot Forecasting}
\label{appendix:zero_shot}
Following the zero-shot forecasting of FPT~\cite{jin2023time}, each experiment comprises the source and target datasets. We train a model on the source dataset and apply the model on the target dataset for predictions directly. 

Notably, the zero-shot scenarios are conducted respectively on the subsets (e.g. M4 Monthly $\to$ M3 Monthly) and the subsets are divided by the sampling frequency but follow different distributions~\cite{makridakis2020m4}. 

For M4 $\to$ M3, which means training on M4 and testing on M3, we directly utilize the same model in the short-term forecasting experiments reported in Table~\ref{tab:forecast_short_full}. Considering different horizons in subsets, for M3 Yearly, M3 Quarterly, and M3 Monthly, we directly employ models trained on corresponding subsets of M4 for testing. As for M3 Others, we test using the model trained on M4 Quarterly to keep the same horizon. 

For M3 $\to$ M4, similarly, for M4 Yearly, M4 Quarterly, and M4 Monthly, we directly employ models trained on corresponding subsets of M3 for testing. For the remaining subsets, M4 Weekly, M4 Daily, and M4 Hourly, we perform inference using the model trained on M3 Monthly. Table~\ref{tab:forecast_zeroshot_full} shows the detailed result.

\begin{table*}[htbp]
  \caption{Results of zero-shot forecasting. We adopt the same protocol as FPT~\cite{zhou2023one}. M4 $\to$ M3 means training forecasters on M4 datasets and evaluating the performance on M3, and vice versa. Results of compared baselines are reported from FPT~\cite{zhou2023one}. Lower SMAPE indicates better performance.}
  \vspace{5pt}
  \label{tab:forecast_zeroshot_full}
  \centering
  \begin{threeparttable}
  \begin{small}
  \renewcommand{\multirowsetup}{\centering}
  \setlength{\tabcolsep}{2pt}
  \resizebox{\textwidth}{!}{\begin{tabular}{c|c|cccccccccccccccc}
    \toprule
    \multicolumn{2}{c|}{Method} & \textbf{AutoTimes} & FPT & DLinear & PatchTST & TimesNet & NSformer & FEDformer & Informer & Reformer  \\ 
    \toprule
    \multirow{5}{*}{\rotatebox{90}{M4$\ \to\ $M3}} 
    & Yearly    & \textbf{15.71} & 16.42 & 17.43 & \underline{15.99} & 18.75 & 17.05 & 16.00 & 19.70 & 16.03\\
    & Quarterly & \textbf{9.35} & 10.13 & 9.74 & 9.62 & 12.26 & 12.56 & \underline{9.48} & 13.00 & 9.76 \\
    & Monthly   & \underline{14.06} & 14.10 & 15.65 & 14.71 & \textbf{14.01} & 16.82 & 15.12 & 15.91 & 14.80 \\
    & Others  & \underline{5.79} & \textbf{4.81} & 6.81 & 9.44 & 6.88 & 8.13 & 8.94 & 13.03 & 7.53 \\
    & Average & \textbf{12.75} & \underline{13.06} & 14.03 & 13.39 & 14.17 & 15.29 & 13.53 & 15.82 & 13.37 \\
    \midrule
    \multirow{5}{*}{\rotatebox{90}{M3$\ \to\ $M4}} 
    & Yearly    & \textbf{13.728} & \underline{13.740} & 14.193 & 13.966 & 15.655 & 14.988 & 13.887 & 18.542 & 15.652\\
    & Quarterly & \textbf{10.742} & \underline{10.787} & 18.856 & 10.929 & 11.877 & 11.686 & 11.513 & 16.907 & 11.051\\
    & Monthly   & \textbf{14.558} & \underline{14.630} & 14.765 & 14.664 & 16.165 & 16.098 & 18.154 & 23.454 & 15.604 \\
    & Others  & \textbf{6.259} & 7.081 & 9.194 & 7.087 & \underline{6.863} & 6.977 & 7.529 & 7.348 & 7.001\\
    & Average & \textbf{13.036} & \underline{13.125} & 15.337 & 13.228 & 14.553 & 14.327 & 15.047 & 19.047 & 14.092\\
    \bottomrule
  \end{tabular}}
  \end{small}
  \end{threeparttable}
\end{table*}

\subsection{Method Generality}
Mainstream LLMs predominantly adopt the decoder-only architecture, and AutoTimes can utilize any decoder-only LLM. 
We conduct experiments on various types and sizes of LLMs, including GPT-2~\cite{radford2019language}, multiple sizes of OPT~\cite{zhang2022opt}, and LLaMA~\cite{touvron2023llama}. Detailed configurations and results are shown in Table~\ref{tab:llm_ablation_config} and~\ref{tab:llm_ablation_full}, demonstrating a general trend where performance improves as the model size increases, consistent with the scaling law~\cite{kaplan2020scaling}. 

\begin{table*}[htbp]
  \caption{Detailed method configurations of AutoTimes for alternative language models.}
  \vspace{-2pt}
  \vskip 0.1in
  \label{tab:llm_ablation_config}
  \footnotesize
  \begin{threeparttable}
  \begin{small}
  \renewcommand{\multirowsetup}{\centering}
  \setlength{\tabcolsep}{3pt}
  \resizebox{\textwidth}{!}{\begin{tabular}{c|ccccccc}
    \toprule
    Base LLM & GPT-2 (124M) & OPT-350M & OPT-1.3B & OPT-2.7B & OPT-6.7B & LLaMA-7B \\ 
    \toprule
    Hidden Dim. & 768	& 1024	& 2048	& 2560	& 4096	& 4096 \\
    \midrule
    Embedding	& 2-layer MLP	& 2-layer MLP	& 2-layer MLP	& 2-layer MLP	& 2-layer MLP	& Linear \\
    \midrule
    Trainable Param. (M)	&0.44	&0.58	&1.10	&1.36	&2.15	&0.79 \\
    \bottomrule
  \end{tabular}}
  \end{small}
  \end{threeparttable}
\end{table*}

\begin{table*}[htbp]
  \caption{Full Results of alternative LLMs, which are adapted with the context length $C = 672$.}
  \vspace{-2pt}
  \vskip 0.1in
  \label{tab:llm_ablation_full}
  \footnotesize
  \begin{threeparttable}
  \begin{small}
  \renewcommand{\multirowsetup}{\centering}
  \setlength{\tabcolsep}{6pt}
  \resizebox{\textwidth}{!}{\begin{tabular}{c|c|cc|cc|cc|cc|cc|cc}
    \toprule
    \multicolumn{2}{c|}{LLM} & \multicolumn{2}{c|}{GPT-2 (124M)} & \multicolumn{2}{c|}{OPT-350M} & \multicolumn{2}{c|}{OPT-1.3B} & \multicolumn{2}{c|}{OPT-2.7B} & \multicolumn{2}{c|}{OPT-6.7B} & \multicolumn{2}{c}{LLaMA-7B} \\ 
    \cmidrule(lr){1-2} \cmidrule(lr){3-4} \cmidrule(lr){5-6} \cmidrule(lr){7-8} \cmidrule(lr){9-10} \cmidrule(lr){11-12} \cmidrule(lr){13-14}
    \multicolumn{2}{c|}{Metric} & MSE & MAE & MSE & MAE & MSE & MAE & MSE & MAE & MSE & MAE & MSE & MAE \\
    \toprule
    \multirow{5}{*}{\rotatebox{90}{ECL}} 
    & 96  & 0.140 & 0.236 & 0.136 & 0.233 & 0.132 & 0.228 & 0.132 & 0.227 & 0.130 & 0.226 & 0.129 & 0.225\\
    & 192 & 0.159 & 0.253 & 0.154 & 0.249 & 0.150 & 0.245 & 0.149 & 0.244 & 0.148 & 0.242  & 0.147 & 0.241\\
    & 336 & 0.177 & 0.270 & 0.171 & 0.267 & 0.167 & 0.262 & 0.167 & 0.262 & 0.165 & 0.260 & 0.162 & 0.258\\
    & 720 & 0.216 & 0.303 & 0.211 & 0.301 & 0.206 & 0.296 & 0.207 & 0.297 & 0.204 & 0.295 & 0.199 & 0.288\\
    \cmidrule(lr){2-14}
    & Avg & 0.173 & 0.266 & 0.168 & 0.263 & 0.164 & 0.258 & 0.164 & 0.258 & 0.162 & 0.256 & 0.159 & 0.253\\
    \midrule
    \multirow{5}{*}{\rotatebox{90}{ETTh1}} 
    & 96  & 0.360 & 0.397 & 0.365 & 0.403 & 0.357 & 0.395 & 0.360 & 0.398 & 0.357 & 0.397 & 0.360 & 0.400\\
    & 192 & 0.391 & 0.419 & 0.395 & 0.423 & 0.389 & 0.417 & 0.389 & 0.419 & 0.386 & 0.417 & 0.388 & 0.419\\
    & 336 & 0.408 & 0.432 & 0.411 & 0.434 & 0.408 & 0.431 & 0.404 & 0.430 & 0.404 & 0.429 & 0.401 & 0.429\\
    & 720 & 0.429 & 0.452 & 0.432 & 0.457 & 0.430 & 0.452 & 0.421 & 0.447 & 0.427 & 0.454 & 0.406 & 0.440\\
    \cmidrule(lr){2-14}
    & Avg & 0.397 & 0.425 & 0.401 & 0.429 & 0.396 & 0.424 & 0.394 & 0.424 & 0.394 & 0.424 & 0.389 & 0.423\\
    \midrule
    \multirow{5}{*}{\rotatebox{90}{Traffic}} 
    & 96  & 0.369 & 0.257 & 0.371 & 0.260 & 0.361 & 0.253 & 0.358 & 0.251 & 0.357 & 0.251 & 0.343 & 0.248\\
    & 192 & 0.394 & 0.268 & 0.393 & 0.270 & 0.383 & 0.263 & 0.380 & 0.261 & 0.379 & 0.261 & 0.362 & 0.257\\
    & 336 & 0.413 & 0.278 & 0.411 & 0.279 & 0.402 & 0.273 & 0.399 & 0.271 & 0.398 & 0.272 & 0.379 & 0.266\\
    & 720 & 0.449 & 0.299 & 0.446 & 0.300 & 0.440 & 0.295 & 0.437 & 0.293 & 0.437 & 0.294 & 0.413 & 0.284\\
    \cmidrule(lr){2-14}
    & Avg & 0.406 & 0.276 & 0.405 & 0.277 & 0.397 & 0.271 & 0.394 & 0.269 & 0.393 & 0.270 & 0.374 & 0.264\\
    \midrule
    \multirow{5}{*}{\rotatebox{90}{Weather}} 
    & 96 & 0.158 & 0.208 & 0.157 & 0.208 & 0.157 & 0.207 & 0.157 & 0.207 & 0.159 & 0.209 & 0.153 & 0.203\\
    & 192 & 0.207 & 0.254 & 0.205 & 0.252 & 0.207 & 0.253 & 0.206 &0.253 & 0.208 & 0.256 & 0.201 & 0.250\\
    & 336 & 0.262 & 0.298 & 0.261 & 0.294 & 0.263 & 0.296 & 0.265 & 0.297 & 0.268 & 0.302 & 0.256 & 0.293\\
    & 720 & 0.342 & 0.353 & 0.335 & 0.346 & 0.334 & 0.347 & 0.344 & 0.351 & 0.354 & 0.360 & 0.235 & 0.273\\
    \cmidrule(lr){2-14}
    & Avg & 0.242 & 0.278 & 0.240 & 0.275 & 0.240 & 0.276 & 0.243 & 0.277 & 0.247 & 0.282 & 0.235 & 0.273\\
    \bottomrule
  \end{tabular}}
  \end{small}
  \end{threeparttable}
\end{table*}

\subsection{Variable Lookback Length}\label{sec:vary_lookback}
\begin{figure}[htbp]
\begin{center}
    \centerline{\includegraphics[width=\textwidth]{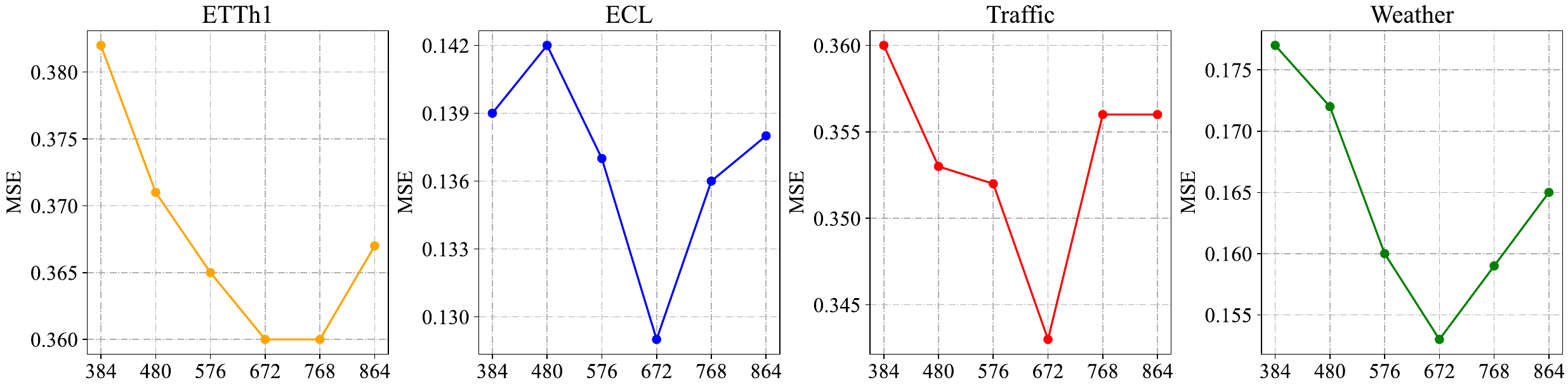}}
    \vspace{0pt}
	\caption{Performance of LLM-based forecasters on the pred-$96$ scenario, which are adapted by AutoTimes by the context length $C=672$ and directly applied on different lookback lengths.}
	\label{fig:vary_lookback}
\end{center}
\vspace{-25pt}
\end{figure}

In the conventional forecasting paradigm, deep forecasters are trained respectively on lookback/forecast lengths, limiting their applicability to a single lookback length. In contrast, LLMs have the versatility to handle various input lengths. This capability is derived from Rotary Position Embedding ~\cite{su2024roformer} and the next toke prediction objective, where LLMs are trained with token-wise supervision in Equation~\ref{equ:loss}, that is, the generated token at each position is supervised. While non-autoregressive LLM4TS methods are typically constrained to a fixed lookback setting, AutoTimes with a consistent training objective has the flexibility to deal with different lookback lengths. We present the results in Figure~\ref{fig:vary_lookback}, where we adapt the LLM by AutoTimes with the context length of $C=672$, and evaluate the performance with different lookback lengths, which demonstrates the inherited versatility of LLM-based forecasters. Moreover, the performance is generally improving with increased available lookback observations, leading to an averaged $9.3\%$ promotion from $384$ to $672$. By contrast, several works have observed that the performance of respectively trained deep forecasters does not necessarily improve with the increasing of lookback length~\cite{liu2023itransformer, nie2022time, zeng2023transformers}.

\subsection{Timestamps as Position Embeddings}\label{sec:timestamp}
\begin{figure*}[htbp]
\begin{center}
    \center{\includegraphics[width=\textwidth]{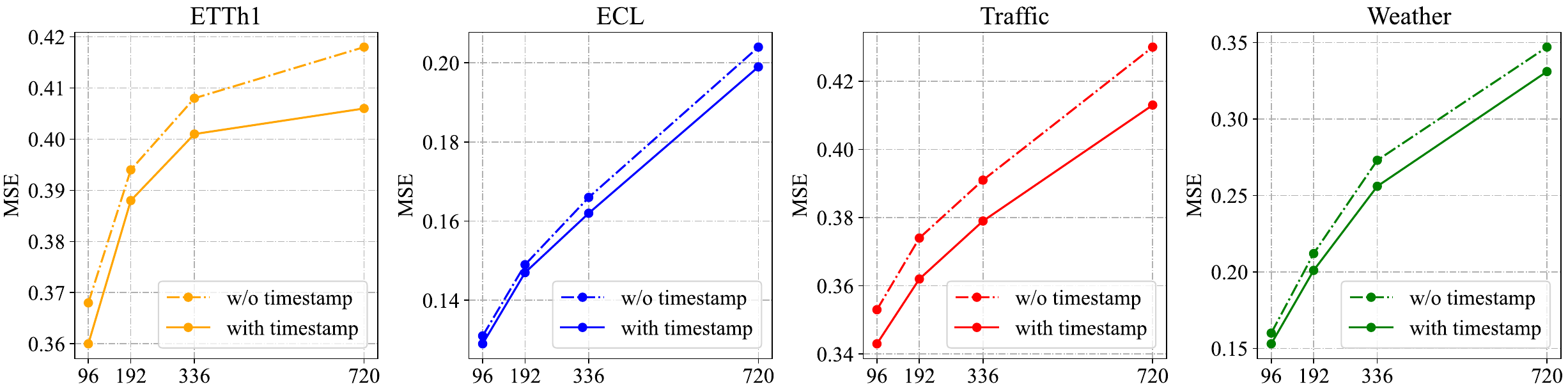}}
	\vspace{-15pt}
    \caption{Ablation on whether to utilize textual timestamps as the position embedding. Results of different prediction lengths are provided, where the embedding leads to consistent performance promotion across all datasets, and the promotion can increase with a longer prediction length.}
	\label{fig:ablation}
\end{center}
\vspace{-10pt}
\end{figure*}

We conduct an ablation study on the proposed position embeddings that integrate timestamps, a prevalent textual covariate in real-world applications. As shown in Figure~\ref{fig:ablation}, the forecasting performance is consistently promoted across all multivariate datasets and prediction lengths. The steady improvement can be attributed to the fact that timestamps demote the absolute position of time series segments on the timeline, explicitly aligning different variates in multivariate scenarios. The increasing promotion with longer prediction length also implies that chronological information, such as date and periodicity, yields significant benefits for long-term forecasting.

\subsection{In-Context Forecasting}\label{sec:icl_detail}
For in-context forecasting, similar to zero-shot forecasting in Appendix~\ref{appendix:zero_shot}, we train our model using the source dataset and directly evaluate it on the target dataset. In this task, we first choose M4 as the source dataset and M3 as the target dataset. It is important to note that the structure of the M3 and M4 datasets differs from typical datasets used for long-term forecasting. They consist of multiple univariate time sequences of different lengths. The final part of each sequence serves as the test set, while the preceding part is used for training.

\paragraph{Implementation} In zero-shot scenarios, we use a sequence of length $F$ preceding and consecutive to the test set as input, referred to as the lookback window, where $F$ is the forecast length of each subset. During in-context forecasting, we concatenate the first $2F$ time points that belong to the same sequence with the lookback window as input. We aim to enhance prediction performance by incorporating more contextual information. Too short sequences ($\le 4F$) are discarded to prevent overlap between the prompt and the lookback window. For a fair comparison, both zero-shot and in-context forecasting performance are reported on the same remaining sequences. Figure~\ref{fig:icl_example} provides showcases of zero-shot and in-context forecasting.

\paragraph{Prompt engineering}
Regarding in-context learning, we further delve into the effect of different strategies to retrieve time series as prompts, which is provided in Table~\ref{tab:prompt_select}. \textbf{P.1} and \textbf{P.2} correspond to the zero-shot and in-context forecasting evaluated in Section~\ref{sec:incontext}. The prompt of \textbf{P.3} contains the last $2F$ time points preceding the beginning of the lookback window. Another retrieval of prompts as \textbf{P.4}, adopts time series that come from another uncorrelated time series (out-of-series). We can obtain the following observations:
\begin{itemize}
    \item \textbf{P.1} \emph{v.s.} \textbf{P.4} indicates that the selected prompt is not suitable, since the prompt does not come from the earlier observations of the same series to be predicted. Although the context window becomes larger, the averaged performance will deteriorate because of irrelevant prompts.
    \item \textbf{P.2} \emph{and} \textbf{P.3} indicates that in most cases, selecting the relevant $2F$ time series from
    the same series can provide better contextual information.
\end{itemize}

This highlights the prompt engineering for in-context forecasting. An intuitive suggestion is to utilize consecutive, inter-periodic, and multiple prompts. To verify this idea, we analyze the periodic effect of time series prompts.

\begin{table}[htbp]
  \caption{Effects of different strategies to retrieve time series as prompts for in-context forecasting.}\label{tab:prompt_select}
  \vspace{-2pt}
  \centering
  \begin{threeparttable}
  \begin{small}
  \renewcommand{\multirowsetup}{\centering}
  \resizebox{\textwidth}{!}{\begin{tabular}{l|ccccc}
    \toprule
    Context for prediction & 
    \rotatebox{0}{\scalebox{1.0}{Yearly}} &
    \rotatebox{0}{\scalebox{1.0}{Quarterly}} &
    \rotatebox{0}{\scalebox{1.0}{Monthly}} &
    \rotatebox{0}{\scalebox{1.0}{Others}} &
    \rotatebox{0}{\scalebox{1.0}{Averaged Err.}} \\
    \toprule
    \scalebox{1.0}{\textbf{P.1}: Lookback $F$} & 21.52 & 12.03 & 13.09 & 8.46 & 13.61 \ \ \\
    \scalebox{1.0}{\textbf{P.2}: Prompt from first $2F$ + Lookback $F$} & 17.03 & 10.29 & 12.24 & 5.33 & 11.80 $\downarrow$ \\
    \scalebox{1.0}{\textbf{P.3}: Prompt from last $2F$ + Lookback $F$} & 16.30 & 9.59 & 12.09 & 6.24 & 11.48 $\downarrow$ \\
    \scalebox{1.0}{\textbf{P.4}: Prompt from $2F$ of other series + Lookback $F$} & 18.95 & 12.18 & 14.37 & 9.46 & 13.98 $\uparrow$ \\
    \bottomrule
  \end{tabular}}
    \end{small}
  \end{threeparttable}
\end{table}

In previous experiments, we adopt M3 and M4 datasets, which are consistent with the zero-shot experiment of FPT~\cite{zhou2023one}, to present the promotion of our in-context paradigm. To provide more rigorous conclusions, we extend the evaluation to widely recognized datasets. Details of the experiment are as follows: By using a trained model checkpoint on a source domain (Traffic), we conduct forecasting without gradient update on target ETT datasets. We evaluate the pred-$96$ performance on the last variate (OT). For the zero-shot scenario, the input is length-$288$ lookback series. For in-context forecasting, the input is (length-$384$ series prompt + length-$288$ lookback series). Considering the dataset periodicity, the prompt is uniformly selected as the Ahead-24 (one-day-ahead) series of the original lookback series. To eliminate the performance boost that comes from extending the input length, we also provide the results of length-$672$ lookback series in the zero-shot scenario. Moreover, we further delve into the effect of different strategies to select time series prompts:

\begin{itemize}
    \item \textbf{Ahead-Period}: The prompt is uniformly selected as the Ahead-$24$ series of the original lookback series where $24$ is one of the periods (daily period) of ETT.
    \item \textbf{Ahead-Random}: The prompt is randomly selected as the previous series of the original  series.
    \item \textbf{Fixed Prompt}: The prompt is fixed as the first $384$ time points in the variate-OT.
    \item \textbf{Other Variate}: The prompt is uniformly selected as Ahead-$24$ series, but comes from other variates.
\end{itemize}

\begin{table}[htbp]
  \caption{Strategies to select time series prompts based on periodicity for in-context forecasting.}\label{tab:prompt_period}
  \vspace{-2pt}
  \centering
  \begin{threeparttable}
  \begin{small}
  \renewcommand{\multirowsetup}{\centering}
  \resizebox{\textwidth}{!}{\begin{tabular}{l|ccccc}
    \toprule
    Context for prediction & 
    \rotatebox{0}{\scalebox{1.0}{ETTh1-OT}} &
    \rotatebox{0}{\scalebox{1.0}{ETTh2-OT}} &
    \rotatebox{0}{\scalebox{1.0}{ETTm1-OT}} &
    \rotatebox{0}{\scalebox{1.0}{ETTm2-OT}} &
    \rotatebox{0}{\scalebox{1.0}{Average Err.}} \\
    \toprule
    \scalebox{1.0}{\textbf{P.0}: Zero-Shot (Input-$288$))} & 0.0673 & 0.1637 & 0.0424 & 0.1669 &  0.1101 \\
    \scalebox{1.0}{\textbf{P.1}: Zero-Shot (Input-$672$)} & 0.0657 & 0.1538 & 0.0415 & 0.1701 &  0.1078 \\
    \scalebox{1.0}{\textbf{P.2}: Ahead-Period (Input-$672$)} & \textbf{0.0645} & \textbf{0.1513} & \textbf{0.0399} & \textbf{0.1629} & \textbf{0.1047} \\
    \scalebox{1.0}{\textbf{P.3}: Ahead-Random (Input-$672$)} & 0.0666 & 0.1621 & 0.0407 & 0.1719 & 0.1103  \\
    \scalebox{1.0}{\textbf{P.4}: Fixed Prompt (Input-$672$)} & 0.0769 & 0.1859 & 0.0512 & 0.2104 &  0.1311 \\
    \scalebox{1.0}{\textbf{P.5}: Other-Variates (Input-$672$)} & 0.1263 & 0.1780 & 0.0852 & 0.2297 &  0.1548 \\
    \bottomrule
  \end{tabular}}
    \end{small}
  \end{threeparttable}
  \vspace{-20pt}
\end{table}

Results in Table~\ref{tab:prompt_period} demonstrate the effectiveness of using suitable time series prompts and highlight the influence of prompt engineering. Using inter-period prompts can outperform simply extending lookback window.

The benefit of the proposed in-context forecasting is to extend the input context of time series forecasting beyond a continuous lookback window. Since the essence of prompts is to incorporate useful domain-specific knowledge, here is one use case of in-context forecasting: Considering predicting the weather of one day, one approach is to extend the lookback length from days to weekends. However, it can also introduce noisy information since non-stationary meteorological conditions can change with seasons. Another practical way is to consider how the weather changes on the same day in the last year (or years). Although the input is not continuous, the input context becomes more relevant based on prior knowledge about the periodicity (yearly). Therefore, in-context forecasting makes prior knowledge incorporatable and gets performance promotion.

\subsection{Ablation Study} In addition to the ablation study of whether LLMs are useful in AutoTimes (Table~\ref{tab:ablation}), we further delve into the main difference between our method and previous LLM4TS approach and provide a comprehensive ablation study. The results presented in Table~\ref{tab:ablation_supp} demonstrate that the performance of non-autoregression projection is consistently inferior to that of our autoregressive AutoTimes approach.

\begin{table}[htbp]
\caption{Ablation study of the autoregression. \emph{FlattenHead} replaces the segment-wise projection of AutoTimes by flatten and linear head~\cite{nie2022time}, which is prevalent in non-autoregressive forecasters.}
  \label{tab:ablation_supp}
  \centering
  \begin{threeparttable}
  \begin{small}
  \renewcommand{\multirowsetup}{\centering}
  \setlength{\tabcolsep}{2pt}
  \resizebox{\textwidth}{!}{\begin{tabular}{l|c|cccc|cccc|cccc|cccc}
    \toprule
    \multicolumn{2}{c|}{Dataset} &  \multicolumn{4}{c|}{ETTh1}  &  \multicolumn{4}{c|}{ECL} &  \multicolumn{4}{c|}{Weather}  &  \multicolumn{4}{c}{Traffic}\\
    \cmidrule(lr){1-2}\cmidrule(lr){3-6}\cmidrule(lr){7-10} \cmidrule(lr){11-14}\cmidrule(lr){15-18} 
    \multicolumn{2}{c|}{Type} & \multicolumn{2}{c}{\scalebox{0.80}{\textbf{AutoTimes}}} &
    \multicolumn{2}{c}{\scalebox{0.80}{FlattenHead}}  & 
    \multicolumn{2}{c}{\scalebox{0.80}{\textbf{AutoTimes}}} &
    \multicolumn{2}{c|}{\scalebox{0.80}{FlattenHead}}  & 
    \multicolumn{2}{c}{\scalebox{0.80}{\textbf{AutoTimes}}} &
    \multicolumn{2}{c}{\scalebox{0.80}{FlattenHead}} & 
    \multicolumn{2}{c}{\scalebox{0.80}{\textbf{AutoTimes}}} & 
 \multicolumn{2}{c}{\scalebox{0.80}{FlattenHead}}  \\
     \cmidrule(lr){1-2}\cmidrule(lr){3-4} \cmidrule(lr){5-6}\cmidrule(lr){7-8} \cmidrule(lr){9-10}\cmidrule(lr){11-12}\cmidrule(lr){13-14}\cmidrule(lr){15-16}\cmidrule(lr){17-18}
    \multicolumn{2}{c|}{Metric} & \scalebox{0.90}{MSE} & \scalebox{0.90}{MAE} & \scalebox{0.90}{MSE} & \scalebox{0.90}{MAE} & \scalebox{0.90}{MSE} & \scalebox{0.90}{MAE} & \scalebox{0.90}{MSE} & \scalebox{0.90}{MAE} & \scalebox{0.90}{MSE} & \scalebox{0.90}{MAE} & \scalebox{0.90}{MSE} & \scalebox{0.90}{MAE} & \scalebox{0.90}{MSE} & \scalebox{0.90}{MAE} & \scalebox{0.90}{MSE} & \scalebox{0.90}{MAE}  \\
    \toprule
    \multicolumn{2}{c|}{\scalebox{0.90}{Pred-$96$}}
    & \scalebox{0.90}{\textbf{0.360}} & \scalebox{0.90}{\textbf{0.400}} 
    & \scalebox{0.90}{0.385} & \scalebox{0.90}{0.420} 
    & \scalebox{0.90}{\textbf{0.129}} & \scalebox{0.90}{\textbf{0.225}}    
    & \scalebox{0.90}{0.142} & \scalebox{0.90}{0.247} 
    & \scalebox{0.90}{\textbf{0.153}} & \scalebox{0.90}{\textbf{0.203}}
    & \scalebox{0.90}{0.155} & \scalebox{0.90}{0.209} 
    & \scalebox{0.90}{\textbf{0.343}} & \scalebox{0.90}{\textbf{0.248}} 
    & \scalebox{0.90}{0.367} & \scalebox{0.90}{0.261} \\
    
    \midrule
    \multicolumn{2}{c|}{\scalebox{0.90}{Pred-$192$}} 
    & \scalebox{0.90}{\textbf{0.388}} & \scalebox{0.90}{\textbf{0.419}} 
    & \scalebox{0.90}{0.445} & \scalebox{0.90}{0.463}
    & \scalebox{0.90}{\textbf{0.147}} & \scalebox{0.90}{\textbf{0.241}} 
    & \scalebox{0.90}{0.157} & \scalebox{0.90}{0.259}
    & \scalebox{0.90}{\textbf{0.201}} & \scalebox{0.90}{\textbf{0.250}}
    & \scalebox{0.90}{0.202} & \scalebox{0.90}{0.251} 
    & \scalebox{0.90}{\textbf{0.362}} & \scalebox{0.90}{\textbf{0.257}} 
    & \scalebox{0.90}{0.391} & \scalebox{0.90}{0.282} \\

    \midrule
    \multicolumn{2}{c|}{\scalebox{0.90}{Pred-$336$}} 
    & \scalebox{0.90}{\textbf{0.401}} & \scalebox{0.90}{\textbf{0.429}} 
    & \scalebox{0.90}{0.463} & \scalebox{0.90}{0.475} 
    & \scalebox{0.90}{\textbf{0.162}} & \scalebox{0.90}{\textbf{0.258}} 
    & \scalebox{0.90}{0.201} & \scalebox{0.90}{0.311}  
    & \scalebox{0.90}{\textbf{0.256}} & \scalebox{0.90}{\textbf{0.293}} 
    & \scalebox{0.90}{0.257} & \scalebox{0.90}{0.286} 
    & \scalebox{0.90}{\textbf{0.379}} & \scalebox{0.90}{\textbf{0.266}} 
    & \scalebox{0.90}{0.404} & \scalebox{0.90}{0.287} \\
    
    \midrule
    \multicolumn{2}{c|}{\scalebox{0.90}{Pred-$720$}} 
    & \scalebox{0.90}{\textbf{0.406}} & \scalebox{0.90}{\textbf{0.440}} 
    & \scalebox{0.90}{0.574} & \scalebox{0.90}{0.542} 
    & \scalebox{0.90}{\textbf{0.199}} & \scalebox{0.90}{\textbf{0.288}} 
    & \scalebox{0.90}{0.232} & \scalebox{0.90}{0.331}
    & \scalebox{0.90}{\textbf{0.331}} & \scalebox{0.90}{\textbf{0.345}}
    & \scalebox{0.90}{0.333} & \scalebox{0.90}{0.261} 
    & \scalebox{0.90}{\textbf{0.413}} & \scalebox{0.90}{\textbf{0.284}} 
    & \scalebox{0.90}{0.432} & \scalebox{0.90}{0.294} \\
    \bottomrule
  \end{tabular}}
  \end{small}
  \end{threeparttable}
  \vspace{-5pt}
\end{table}

\section{Showcases}\label{sec:showcases}
\begin{figure}[htbp]
\begin{center}
    \centerline{\includegraphics[width=1\columnwidth]{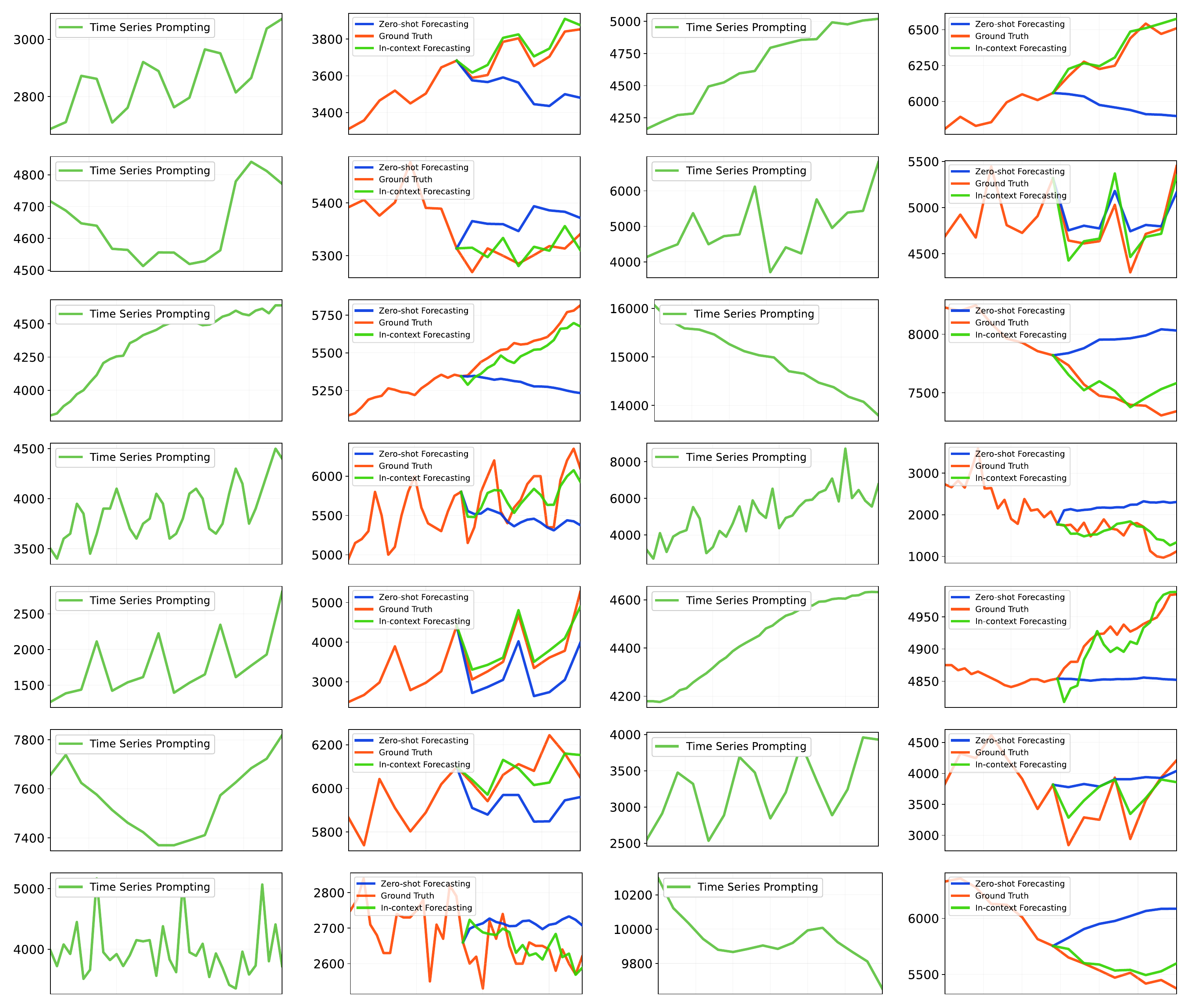}}
    \vspace{-5pt}
	\caption{Showcases of zero-shot and in-context forecasting. For in-context forecasting, beyond the lookback window, we uniformly adopt the first $2F$ time points that belong to the same sequence as the prompt and concatenate them as the prediction context, which achieves a more accurate prediction.}
	\label{fig:icl_example}
\end{center}
\end{figure}

To facilitate a clear comparison among various models, we present additional prediction showcases for long-term forecasting and short-term forecasting. These examples are provided by the following models: TimeLLM~\cite{jin2023time}, FPT~\cite{zhou2023one} and PatchTST~\cite{nie2022time}. Of all the models, AutoTimes delivers the most accurate future series predictions. Additionally, we provide the showcases of zero-shot and in-context forecasting in Figure~\ref{fig:icl_example}.

\begin{figure}[ht]
\begin{center}
    \centerline{\includegraphics[width=\columnwidth]{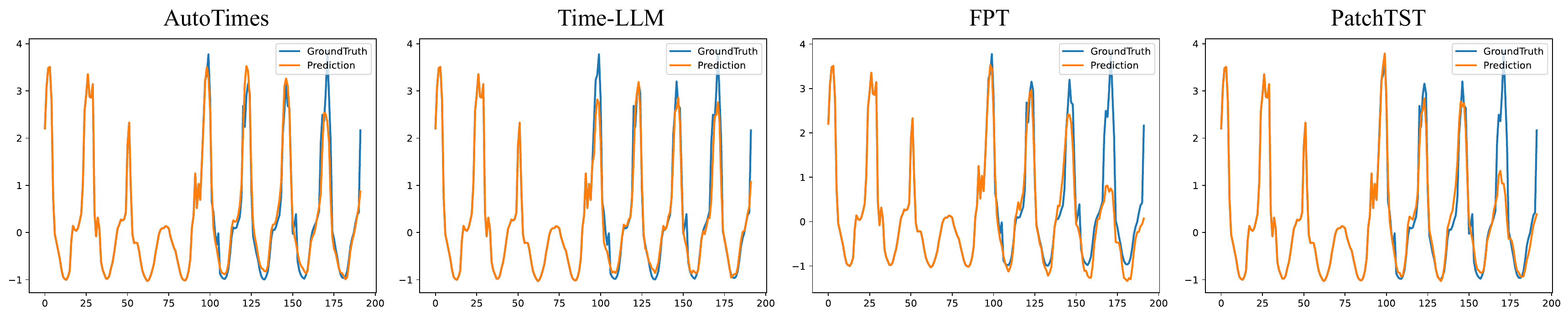}}
    \vspace{-5pt}
	\caption{Visualization of input-$672$-predict-$96$ results on the Traffic dataset.}
	\label{fig:long_term_showcase}
\end{center}
\vspace{-40pt}
\end{figure}

\begin{figure}[ht]
\begin{center}
    \centerline{\includegraphics[width=\columnwidth]{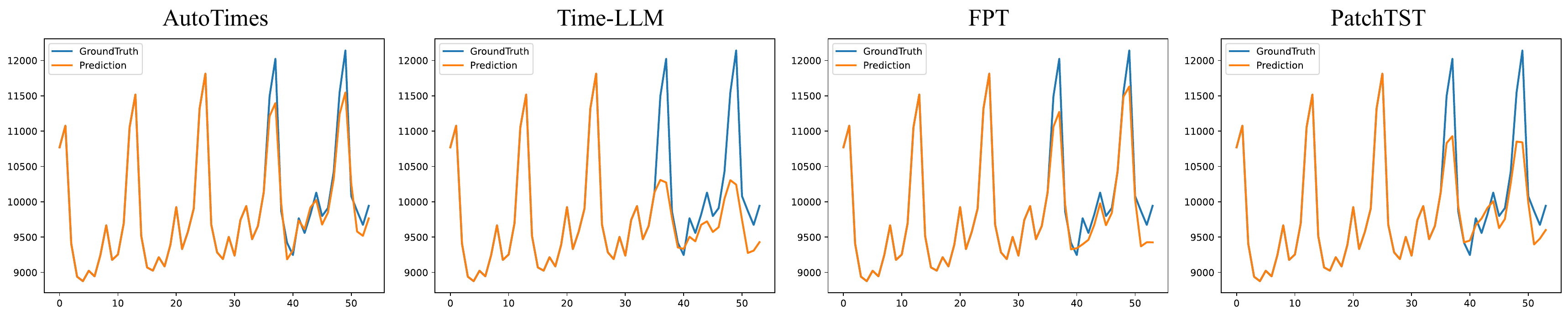}}
    \vspace{-5pt}
	\caption{Visualization of input-$36$-predict-$18$ results on the M4 Monthly dataset.}
	\label{fig:short_term_showcase}
     \vspace{-10pt}
\end{center}
\end{figure}

\newpage
\section{Broader Impact}\label{sec:board}
\subsection{Impact on Real-world Applications}
This paper copes with general-purpose time series forecasting, which is faced with increasing challenges such as the versatility to handle variable-length scenarios, good generalizability with scarce samples, utilization of multimodality, and instructive downstream prompts. Since previous studies have demonstrated the feasibility of leveraging large language models for time series, we propose a simple but effective approach as AutoTimes to obtain LLM-based forecasters, which keeps the consistency of autoregression. Our model achieves state-of-the-art performance on forecasting benchmarks and demonstrates remarkable adaptation speed and parameter efficiency. Besides, advanced capabilities such as multi-step generation and in-context learning are inherited by the repurposed forecaster. Therefore, the proposed method makes it promising to tackle real-world applications, which helps our society prevent risks in advance and make better decisions with limited computational budgets. Our paper mainly focuses on scientific research and has no obvious negative social impact.

\subsection{Impact on Future Research}
In this paper, we find prevalent non-autoregressive LLM4TS methods have inconsistencies in the model structure and generative approach with LLMs, leading to insufficient utilization of the inherent multi-step token transition. Given that the generalizability and generative ability of LLMs are largely derived from the autoregressive manner, the potentials of LLMs may not be fully exhibited in time series forecasting. Therefore, we propose to adapt LLMs by the consistent training objective, the next token prediction, and accomplish arbitrary-length forecasting by iterative generation. Beyond the conventional forecasting paradigm, we propose in-context forecasting, where the context for prediction is extended, and earlier historical time series can be utilized as advantageous prompts. The compatibility with LLMs and insights from autoregression can be instructive for future LLM4TS research and the development of foundation time series models.

\section{Limitation}\label{sec:limitation}
The proposed method has not supported probabilistic forecasting, since AutoTimes only establishes the mapping between time series segments to latent embeddings of the LLM, instead of discrete language tokens. Advanced low-rank adaptation is under exploration in our work, which can further align suitable token transitions as the future extrapolation of time series. More deftly designed embedding and projection layers are underexplored to support more compatible tokenization for time series. Besides, it is fascinating to apply AutoTimes on real-world multimodal time series datasets (such as news-stocks, and logs-measurements), which leaves our future work.

\end{document}